\documentclass[runningheads]{llncs}

\usepackage{eccv}

\usepackage{eccvabbrv}

\usepackage{graphicx}
\usepackage{booktabs}

\usepackage[accsupp]{axessibility}  

\usepackage{hyperref}

\usepackage{orcidlink}
\usepackage{algorithm}
\usepackage{algorithmic}
\usepackage{amsmath}
\usepackage{amssymb}
\usepackage{tabularx}
\usepackage{booktabs}
\usepackage{multirow}
\usepackage{graphicx}
\usepackage[normalem]{ulem}
\useunder{\uline}{\ul}{}
\usepackage{enumitem}
\usepackage{rotating}
\usepackage{adjustbox}
\usepackage{array}     

\begin{document}

\title{Holistic Optimal Label Selection for Robust Prompt Learning under Partial Labels} 

\titlerunning{HopS}


\author{Yaqi Zhao\orcidlink{0009-0005-0390-1968} \and Haoliang Sun\thanks{Corresponding author.}\orcidlink{0000-0001-7715-5682} \and Yating Wang\orcidlink{0009-0002-8687-0471} \and Yongshun Gong\orcidlink{0000-0003-3948-4471} \and Yilong Yin\orcidlink{0000-0002-8465-1294}}

\authorrunning{Y.~Zhao et al.}


\institute{School of Software, Shandong University, Jinan, China }
\maketitle

\begin{abstract}
  Prompt learning has gained significant attention as a parameter-efficient approach for adapting large pre-trained vision-language models to downstream tasks. However, when only partial labels are available, its performance is often limited by label ambiguity and insufficient supervisory information. To address this issue, we propose \textit{Holistic Optimal Label Selection} (\textbf{HopS}), leveraging the generalization ability of pre-trained feature encoders through two complementary strategies. First, we design a \textit{local} density-based filter that selects the top frequent labels from the nearest neighbors’ candidate sets and uses the softmax scores to identify the most plausible label, capturing structural regularities in the feature space. Second, we introduce a \textit{global} selection objective based on optimal transport that maps the uniform sampling distribution to the candidate label distributions across a batch. By minimizing the expected transport cost, it can determine the most likely label assignments. These two strategies work together to provide robust label selection from both local and global perspectives. Extensive experiments on eight benchmark datasets show that HopS consistently improves performance under partial supervision and outperforms all baselines. Those results highlight the merit of holistic label selection and offer a practical solution for prompt learning in weakly supervised settings. The code is available at https://github.com/Qizhoay/HopS.
  \keywords{Partial label learning \and Prompt learning \and Optimal transport}
\end{abstract}  

\section{Introduction}
\label{sec:intro}
Large pre-trained vision-language models (VLMs), such as CLIP \cite{radford2021learning}, have demonstrated remarkable capabilities across a wide range of downstream tasks~\cite{rombach2022high, frans2022clipdraw, luo2023lexlip, sun2024alpha}. Among various fine-tuning paradigms~\cite{liu2023pre}, prompt tuning has emerged as an efficient alternative to full model fine-tuning, offering the advantage of adapting large models with minimal additional parameters~\cite{zhou2022learning}. By optimizing a small set of learnable prompts while keeping the backbone frozen, prompt learning preserves the generalization strength of the pre-trained model and reduces training costs—making it particularly attractive in resource-constrained or few-shot learning scenarios~\cite{zeng2025local}.

The effectiveness of prompt learning can be significantly compromised in weakly supervised settings~\cite{zhang2024noise}, particularly those involving partial labels~\cite{lv2024makes}, where only a subset of candidate labels is provided per instance without explicit ground-truth annotations. This setting is common in real-world applications such as webly supervised learning~\cite{qin2023capro}, human-in-the-loop labeling~\cite{NEURIPS2024_72de9826}, and open-world recognition~\cite{wen2024cross}. The main challenge in such scenarios lies in the label ambiguity, which hampers supervision and often degenerate the generalization performance of learning algorithms, especially for prompt learning with few-shot instances~\cite{zhou2024few}.

To address the challenges of partial label learning (PLL), state-of-the-art methods typically select a plausible label from the candidate set using representations learned through contrastive learning (e.g., PICO~\cite{wang2022pico}). However, such approaches are not directly applicable to pre-trained models, as fine-tuning their vision encoders may compromise their zero-shot capabilities~\cite{kumar2022fine}. \textit{This presents a fundamental challenge in filling the gap between reliable label selection with frozen vision encoders and effective prompt learning with label disambiguation}.

To overcome this limitation, we propose a holistic label selection strategy that fully leverages the representational generalization of pre-trained encoders for robust prompt learning. Specifically, HopS includes two complementary selection mechanisms. The first is a \textbf{Local Density-based Filter} (LDF), which estimates label frequency within a $k$-nearest neighbor ($k$-NN) structure in the image feature space and selects the most frequent labels in the neighborhood for a subset, capturing local semantic regularities in a non-parametric manner. LDF effectively exploits the zero-shot capabilities of the encoder and avoids overfitting to noisy labels during the early training stages. Among the subset, the most plausible label is then identified based on the softmax scores. The second is a \textbf{Global Optimal Transport Planner} (GOP). It maps a uniform global prior—representing the overall underlying distribution—to the candidate label distributions across a mini-batch. By minimizing the expected transport cost, GOP encourages globally optimal label assignments. The resulting transport plan explicitly characterizes how each instance contributes to the class-wise probability mass in the label distribution, thereby providing a principled and interpretable foundation for identifying and selecting the most relevant label.


By integrating both \textit{local} and \textit{global} perspectives, our framework enables more accurate and stable label selection, mitigating the impact of label ambiguity. We evaluate our method across a variety of vision-language benchmarks under partial supervision and demonstrate consistent improvements over strong baselines. Our key contributions are as follows:
\begin{itemize}
    \item We provide a comprehensive investigation into leveraging the representational generalization of pre-trained VLMs for prompt learning in the partial supervision setting.
    \item We propose a novel holistic label selection framework that combines a local density-based filter with a global optimal transport planner.
    \item We empirically validate the effectiveness of our approach, achieving state-of-the-art performance on multiple benchmarks.
\end{itemize}
\section{Related Work}
\subsection{Partial Label Learning}
PLL \cite{jin2002learning, cour2011learning, chen2014ambiguously, yu2016maximum} assumes that each training instance is associated with a candidate label set, within which only one label is correct but not explicitly specified. To address the ambiguity introduced by these noisy candidates, subsequent research has proposed various strategies, including consistency regularization \cite{feng2020provably,lv2020progressive, wu2022revisiting} and contrastive learning \cite{wang2022pico}. These methods aim to leverage partial supervision to learn more discriminative representations and improve generalization. Expanding on this direction, \cite{NEURIPS2023_6b97236d} introduced a dissimilarity propagation-guided label shrinkage method to refine candidate label sets by eliminating irrelevant labels, thereby enhancing supervision quality. Meanwhile, works such as \cite{xu2021instance, liu2025mixed} have begun to explore the more challenging instance-dependent PLL setting, where candidate label sets are generated based on the characteristics of each instance. More recently, \cite{xia2023towards} and \cite{tian2024crosel} have continued to leverage candidate labels to learn robust representations. Additionally, \cite{lv2024makes} highlighted the critical role of feature representations and label denoising in effective PLL. A related work, SoLar \cite{wang2022solar}, addresses partial label learning under class imbalance by employing OT to align the estimated long-tailed class prior with a uniform distribution. In contrast, our method leverages OT to align the overall underlying label distribution with the candidate label distributions. Furthermore, unlike conventional PLL approaches, our work explores prompt learning for pre-trained VLMs, which leverage powerful vision encoders with strong zero-shot capabilities, rather than training encoders from scratch. 

\subsection{Prompt Learning}
Prompt learning has emerged as a parameter-efficient paradigm for adapting large-scale pre-trained models to downstream tasks~\cite{guo-etal-2025-parameter}. Rather than fine-tuning the entire model, it introduces auxiliary input prompts—either manually crafted templates or learnable continuous embeddings—to guide model predictions. In the vision-language domain, CoOp~\cite{zhou2022learning} first demonstrated that learnable context vectors could serve as effective prompts for adapting CLIP-like models, highlighting their strong transferability. Since then, several studies have enhanced the robustness of prompt learning through strategies such as aligning sample representations in multimodal contexts \cite{tsimpoukelli2021multimodal, chen2022plot, wu2024controlmllm}, employing mixture-of-expert prompts \cite{wang2024one}, and utilizing unsupervised prompt distillation \cite{jin2022unsupervised, li2024cvpr}. In addition, prompt learning has achieved significant improvements on various downstream tasks, including text-to-image generation~\cite{teo2024fairqueue} and open-vocabulary semantic segmentation~\cite{li2024relationship}. While these advances have shown promise in fully supervised settings, their effectiveness under partial supervision remains unexplored. To address this gap, we extend robust prompt tuning~\cite{wu2023prompt, guo2024joapr, pan2025nlprompt} to the PLL scenario, aiming to bridge weak supervision with prompt-based adaptation in pre-trained VLMs. 
\section{Preliminary}
\subsection{Candidate Label Sets in PLL}
We define the training dataset as \(\mathcal{D} = \{(\mathbf{x}_i, S_i)\}_{i=1}^n\), where \(\mathbf{x}_i \in \mathcal{X}\) is an input instance and \(S_i \subseteq \{1, 2, \dots, C\}\) is the candidate label set for a C-category classification task. Here, all elements $\{s_{i1}, \dots, s_{il}\dots, s_{iL}\}$ in $S_i$ contain one ground-truth label and the rest \(L-1\) elements are the confused labels, involving false positive labels for the current instance. Different levels of label confusion is corresponding to the number of confused labels in each candidate set. 

\subsection{Prompt Optimization with The Candidate Set}
Prompt optimization (PO) in CLIP replaces manual prompt engineering by learning continuous context vectors in an end-to-end manner, while keeping the pre-trained model parameters frozen to fully leverage the knowledge encoded within them. Let $V$, $T$ and $\mathbf{t}$ denote the vision encoder, text encoder, and learnable prompt parameters, respectively. $p(s_{ij} \mid \mathbf{x}_i; \mathbf{t})$ denotes the predicted probability of the j-th label within the candidate set $S_i$, computed via the cosine similarity between the visual and textual embeddings. These embeddings are derived from the visual encoder V, and the text encoder T with the input of the prompt vector $\mathbf{t}$ and the class name embedding $\mathbf{v}_j$. For PO under partial-label supervision, the standard cross-entropy loss is computed over the candidate label set $S_i$ as follows:
\begin{equation}
\begin{aligned}
\mathcal{L}_{\text{CE}}&(\mathbf{x}_i, S_i; \mathbf{t}) = \sum_{j=1}^{C} -s_{i,j} \log \left(p(s_{ij} \mid \mathbf{x}_i; \mathbf{t})\right) \\
\text{s.t.} &\quad \sum_{j \in S_i} s_{i,j} = 1, \;\text{and} \; s_{i,j} = 0 \text{ for } j \notin S_i.
\end{aligned}
\end{equation}
The prediction probability of the input $\mathbf{x}_i$ is given by:
\begin{equation}
\label{CoOp}
p(s_{ij} \mid \mathbf{x}_i; \mathbf{t}) = \frac{\exp\left(\cos\left(V(\mathbf{x}_i), T(\mathbf{t}, \mathbf{v}_j)\right)\right)}{\sum_{q=1}^{C} \exp\left(\cos\left(V(\mathbf{x}_i), T(\mathbf{t}, \mathbf{v}_q)\right)\right)}.
\end{equation}

Despite the expressive power of PO, the inherent label ambiguity within each candidate set $S_i$ creates a fundamental gap between partial label learning and fully supervised learning. This introduces two core challenges: (1) effectively disambiguating the candidate labels, and (2) accurately identifying the ground-truth label for each instance.

\section{Methodology}
To identify the ground-truth label and guide the model in learning effective prompt vectors, we propose a holistic optimal label selection strategy for prompt learning in VLMs. HopS integrates both local and global perspectives through two elaborated components: a local density-based filter, which selects the “local-consensus” candidate, and a global optimal transport planner, which identifies the “global-harmony” candidate. These components work in concert to robust and complementary label guidance for effective prompt optimization under partial supervision.

\subsection{Local Density-Based Filter}
The local density of an instance in the image feature space reflects the intrinsic semantic regularities associated with its candidate labels. To leverage this density information for label selection, we identify the $k$-nearest neighbors of each instance $(\mathbf{x}_i, S_i)$ in the image feature space, denoted as $\{(\mathbf{x}_j, \hat{S}_j)\}_{j=1}^k$. Specifically, we build an affinity matrix $\mathcal{A}$ by computing the cosine similarity between image features of all instances, where the features are extracted using the frozen vision encoder of CLIP. Based on $\mathcal{A}$, we retrieve the top-$k$ most similar examples for each instance, thereby capturing local semantic structure. Once the $k$-nearest neighbors are selected, we then construct a multiset-union candidate set for each instance as:
\begin{equation}
	\mathcal{N}_i = \left( \biguplus_{j=1}^k \hat{S}_j \right)  \biguplus S_i,
\label{eq:nei}
\end{equation}
where $ \uplus$ denotes multiset union, preserving label multiplicities. It is worth noting that, since the affinity matrix is pre-computed prior to training, the retrieval step for each instance reduces to a neighbor search with time complexity $\mathcal{O}(n \cdot \log k)$. Given the small number $n$ of training samples in prompt learning, the retrieval cost is negligible.

Next, we compute the frequency of each category $c$ in $\mathcal{N}_i$. Let $f(c)$ denote the relative frequency of element $c$ in the multiset $\mathcal{N}_i$, defined as:
\begin{equation}
	f(c) = \frac{|\{l \mid s_{jl} = c\}|}{|\mathcal{N}_i|},
\label{eq:fc}
\end{equation}
where $|\{ l \mid s_{jl} = c \}|$ counts the number of times element $c$ appears in $ \mathcal{N}_i$, and $|\mathcal{N}_i|$ is the total number of elements in the multiset, including duplicates.

To enforce label consistency and suppress unreliable candidates, we retain categories whose frequency exceeds a predefined threshold $\tau \in [0, 1]$, thereby forming a consensus-aware mask set for each instance as:
\begin{equation}
	\mathcal{M}_i = \{ c \in \mathcal{N}_i \mid f(c) \geq \tau \}.
\label{eq:mi}
\end{equation}
The refined candidate set is then obtained by intersecting the mask set $\mathcal{M}_i$ with the original candidate set $S_i$, i.e., $\mathcal{M}_i \cap S_i$, ensuring that only labels supported by both the instance and its neighbors are retained. To prevent degenerate cases where all candidate labels are eliminated, we revert to the original candidate set $S_i$ if the refined set becomes empty.

Finally, the most plausible candidate label $y^{\text{local}}$ is selected as the one with the highest prediction probability in the refined candidate set, as determined by Eq.~(\ref{CoOp}).

\subsection{Global Optimal Transport Planner}
\begin{figure}[tb]
  \centering
  \includegraphics[width=0.9\columnwidth]{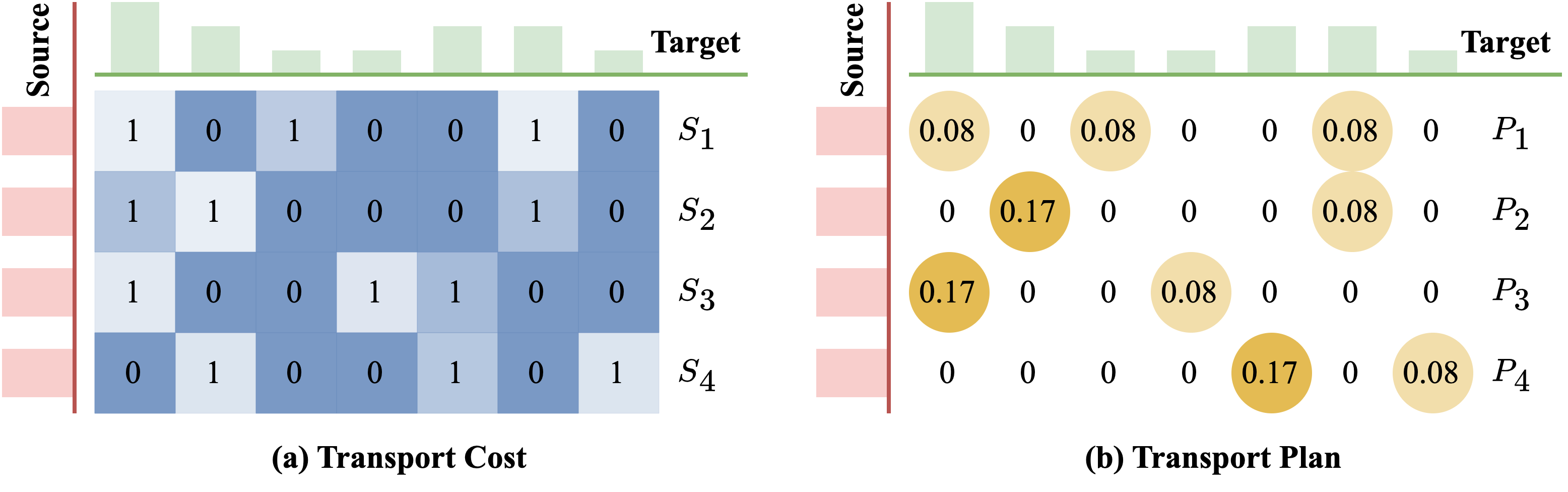}
  \caption{Illustration of the transport cost matrix and the resulting transport plan between the \textbf{uniform source instance distribution} (Source) and the \textbf{target candidate label distribution} (Target).}
  \label{fig:cost}
\end{figure}

To complement the local selection, we introduce a globally consistent label selection mechanism via optimal transport (OT). This formulation matches instances with their candidate labels in a way that aligns with both a uniform prior over instances and the empirical label distribution within the batch, selecting for each instance the label that receives the highest transport mass.
As shown in Fig. \ref{fig:cost}, the matrix on the left illustrates the candidate label sets for four instances in a seven-class classification task, with lighter colors indicating lower transport costs and, consequently, higher label credibility. The right matrix shows the resulting optimal transport plan under the given cost constraints. Here, darker colors indicate greater transported mass, reflecting the planner’s preference for allocating probability mass to more credible classes. The transport process strictly adheres to the marginal constraints, ensuring the consistency of source and target distributions before and after transport.

\noindent\textbf{Discrete Transport Formulation.}
Given a batch of \(B\) samples, we define a uniform source distribution over instences, where $\mathbf{r} \in \Delta^B$ is a $B$-dimensional probability simplex. A candidate-aware marginal distribution is estimated based on candidate frequencies, denoting as $\mathbf{c} \in \Delta^C$. The element in $\mathbf{r}$ and $\mathbf{c}$ are computed as:
\begin{equation}
	r_i = \frac{1}{B}, \quad c_l = \frac{1}{B} \sum_{i=1}^B \frac{s_{il}}{|S_i|}.
\label{eq:rc}
\end{equation}

The transport cost matrix \(\mathbf{M}_{\text{cost}} \in \mathbb{R}^{B \times C}\) in OT is constructed from the similarity between visual and textual representations. This cost formulation encourages semantic alignment between image-label pairs, while discouraging assignments to non-candidate classes by assigning them infinite cost. Specifically, given an instance \(\mathbf{x}_i\), the cost of assigning it to class $j$ is defined as:
\begin{equation}
\mathbf{M}_{\text{cost}}[i, j] = \begin{cases} 1 - \frac{\exp\left(\cos\left(V(\mathbf{x}_i), T(\mathbf{t}, \mathbf{v}_j)\right)\right)}{\sum_{q=1}^{C} \exp\left(\cos\left(V(\mathbf{x}_i), T(\mathbf{t}, \mathbf{v}_q)\right)\right)}, & \text{if } j \in \mathcal{S}_i \\ \infty, & \text{otherwise.} \end{cases}
\label{eq:mcost}
\end{equation}

The optimal transport plan $\mathbf{P}$ aims to move probability mass from a uniform distribution over instances to a candidate-aware label distribution with minimal total cost. To ensure differentiability and enable efficient computation, we incorporate an entropy regularization term $H(\mathbf{P}) = -\sum_{i=1}^{B} \sum_{j=1}^{C} P_{ij} \log P_{ij}$ following \cite{cuturi2013sinkhorn}. The resulting optimal transport objective is formulated as:
\begin{equation}
\label{eq:obj}
\begin{aligned}
    \min_{\mathbf{P} \in \mathbb{R}_+^{B \times C}} \quad \langle \mathbf{P}, \mathbf{M}_{\text{cost}} \rangle - \varepsilon H(\mathbf{P}) \quad \quad \quad\\
    \text{s.t.} \quad \mathbf{P} \mathbf{1}_C = \mathbf{r}, \; \mathbf{P}^\top \mathbf{1}_B = \mathbf{c}, \; P_{ij} = 0 \; \text{if} \; s_{ij} = 0.
\end{aligned}
\end{equation}
Here, $\varepsilon$ is a hyper-parameter controlling the strength of the entropy regularization, and $ \mathbf{1}_C$ and $\mathbf{1}_B $ denote all-one vectors of length $C$ and $B$, respectively.

\noindent\textbf{Approximation Scheme.}
To efficiently solve the entropy-regularized optimal transport problem described in Eq. (\ref{eq:obj}) and compute the transport plan, we adopt the well-known Sinkhorn-Knopp algorithm, an efficient iterative method.  
\begin{equation}
\mathbf{P}^{\text{ot}} = \text{diag}(\mathbf{\boldsymbol{\alpha}}) \cdot \mathbf{\exp\left(-\frac{\mathbf{M}_{\text{cost}}}{\varepsilon} \right)} \cdot \text{diag}(\mathbf{\boldsymbol{\beta} }), 
\label{eq:potsink}
\end{equation}
The iterative steps are given as follows:
\begin{equation}
	\boldsymbol{\alpha} \leftarrow \mathbf{r} \oslash (\mathbf{M} \boldsymbol{\beta}), \quad \boldsymbol{\beta} \leftarrow \mathbf{c} \oslash (\mathbf{M}^\top \boldsymbol{\alpha}).
\label{eq:sinkiter}
\end{equation}
where \(\boldsymbol{\alpha}\) and \(\boldsymbol{\beta}\) are scaling vectors, \(\oslash\) denotes element-wise division, \(\mathbf{r}\) and \(\mathbf{c}\) are the source and target marginal distributions defined above, and \(\mathbf{M}\) is the Gibbs kernel \cite{cuturi2013sinkhorn} derived from the cost matrix.




After computing the optimal transport plan \(\mathbf{P}^{\text{ot}} \in \mathbb{R}^{B \times C}\), which encodes the soft assignment between instances and candidate labels, we select $y^{\text{global}}$ as the label with the largest transport mass. Unlike $y^{\text{local}}$, this global selection considers not only the candidate constraint but also the semantic alignment captured by the transport plan, thereby improving consistency across training instances.

\subsection{Learning Objective and Procedure}
\textbf{The Objective Function.} After select the two most plausible labels $y^{\text{local}}$ and $y^{\text{global}}$, we conduct prompt optimization and jointly optimize the cross-entropy loss. Given an input image \(\mathbf{x}\), the overall loss is defined as:
\begin{equation}
	\label{eq:finalobj}
\arg \min_{\mathbf{t}} \ \mathcal{L}_\text{{\tiny CE}}(p(y \mid \mathbf{x}; \mathbf{t}), y^{\text{local}}) + \lambda \mathcal{L}_\text{{\tiny CE}}(p(y \mid \mathbf{x}; \mathbf{t}), y^{\text{global}}),
\end{equation}
where \(\lambda\) is the weighting coefficient for the two loss components, the prediction probability $p(y \mid \mathbf{x}; \mathbf{t})$ for each category is computed by Eq. (\ref{CoOp}).

\noindent\textbf{Training Procedure.} As illustrated in Algorithm \ref{HopS-code}, the HopS performs both LDF and GOP selection to iteratively refine predictions under partial supervision, resulting in a more robust and discriminative prompt. The computational overhead is negligible (see \cref{tab:time_comparison}).

\begin{algorithm}[t]
\caption{The Holistic Optimal Selection}
\label{HopS-code}
\textbf{Input}: Training dataset \(\mathcal{D} = \{(\mathbf{x}_i, S_i)\}_{i=1}^n\), pre-trained encoders $V$ and $T$\\
\textbf{Parameter}: $k, \tau, \varepsilon, \lambda$\\
\textbf{Output}: The prompt vector $\mathbf{t}$
\begin{algorithmic}[1] 
\STATE Extract and store all image features $ R_\text{vis} \gets V(X)$.
\STATE Calculate $\mathcal{A}$ between $ R_\text{vis}$ by cosine similarity.
\FOR{$batch = 1, 2, \ldots$}
    \STATE \texttt{// LDF}
    \STATE Choose $k$ neighbors to compute $\mathcal{N}_i$ corresponding to each instance by Eq. (\ref{eq:nei}).
    \STATE Calculate the frequency $f(\cdot)$ by Eq. (\ref{eq:fc}).
    \STATE Construct the consensus-aware mask set and refined candidate set by Eq. (\ref{eq:mi}).
    \STATE Identify $Y^{local}$ with the highest prediction probability in the refined candidate set.
    \STATE \texttt{// GOP}
    \STATE Initialize the distribution of $\mathbf{r}$ and $\mathbf{c}$ by Eq. (\ref{eq:rc}).
    \STATE Calculate the cost matrix $\mathbf{M}_\text{cost}$ with $T(t, v_j)$ and $R_\text{vis}[\text{instance indexes}]$ by Eq. (\ref{eq:mcost}).
    \FOR{$iter = 1, 2, \ldots$}
        \STATE \texttt{// Sinkhorn-Knopp Algorithm}
        \STATE Approximate transport plan by Eq. (\ref{eq:potsink}).
    \ENDFOR
    \STATE Gain the $Y^{global}$ with the largest transport mass.
    \STATE Update $\mathbf{t}$ with Eq. (\ref{eq:finalobj}).
\ENDFOR
\end{algorithmic}
\end{algorithm}


\section{Experiments}
\label{sec:Experiments}
We conduct experiments on eight datasets, comparing our approach with eight representative loss functions, zero-shot learning, and 16-shot fully supervised prompt tuning using CoOp. All methods are evaluated under varying degrees of label ambiguity. Our method consistently outperforms all baselines, highlighting its effectiveness in prompt learning with partial labels. Extensive experiments and analyses further demonstrate the complementarity of the two label selection strategies.

\subsection{Experimental Settings}
Basic settings are outlined. Additional details (e.g., hyperparameters) are in the Appendix \cref{sec:exp_set}.

\noindent\textbf{Dataset}.
We adopt eight datasets: Caltech\cite{fei2004learning}, DTD\cite{cimpoi2014describing}, EuroSAT\cite{helber2019eurosat}, FGVCAircraft\cite{maji2013fine}, Food\cite{nilsback2008automated}, Flowers\cite{nilsback2008automated}, OxfordPets\cite{parkhi2012cats}, and UCF\cite{soomro2012ucf101}. These datasets encompass a wide spectrum of visual recognition tasks, especially for challenging fine-grained classification.

\noindent\textbf{Confusion Types}. We manually corrupt the datasets into partially labeled versions using two strategies: random-uniform (\textit{rand}) and instance-dependent (\textit{insd}). (1) In the \textit{rand} setting, confusing labels are randomly selected into the candidate set with equal probability for each instance. (2) The \textit{insd} setting depends on instance-level information. Following \cite{xu2021instance}, we adopt a prototype-based label confusion strategy, which uses pre-trained image encoders to identify the top-\((L-1)\) most similar classes—excluding the ground-truth label—as candidate labels. We employ three visual backbones (i.e., ResNet-18/50 \cite{he2016deep} trained on ImageNet and CLIP-ResNet50 \cite{radford2021learning}) to simulate annotators with varying levels of domain expertise. 

\noindent\textbf{Confusion Levels}. The confusion rate, defined as the ratio of the number of confused labels to \(L\), serves as an indicator of the difficulty in identifying the ground-truth label within the candidate set. The \textit{insd}-confusion and \textit{rand}-confusion levels of the candidate labels are controlled by the size of the candidate label set \(L\), selected from \(\{2, 3, 4, 5, 8, 9, 10\}\), with the corresponding confusion rates $\gamma_c$ being \(\{0.50, 0.67, 0.75, 0.80, 0.88, 0.89, 0.90\}\).

\noindent\textbf{Prompt Types}. The design of prompts can be categorized into two forms, consistent with CoOp: unified prompt (\textit{uni}) and classified prompts (\textit{cls}). The \textit{uni} prompt shares a common set of context vectors across all classes, making it suitable for general classification tasks. In contrast, \textit{cls} assigns an independent set of context vectors to each class. 

\noindent\textbf{Baselines}. We compare HopS with five state-of-the-art partial label learning losses: CC \cite{feng2020provably}, RC \cite{feng2020provably}, LWC \cite{wen2021leveraged}, MSE \cite{feng2020learning}, and EXP \cite{feng2020learning}. Furthermore, we include three well-known robust learning losses, MAE \cite{wang2019symmetric}, SCE \cite{wang2019symmetric}, and GCE \cite{zhang2018generalized}, as well as three recent partial label learning methods, Papi \cite{xia2023towards}, CroSel \cite{tian2024crosel}, and SoLar \cite{wang2022solar}, representing recent advances in the field. All hyperparameters for these methods are carefully tuned following the settings reported in their original publications. The detailed configurations of all compared methods are provided in Appendix Tab.~\ref{tab:methods}.

\noindent\textbf{The Learning Framework}. To ensure a fair comparison, we adopt the most influential context optimization method, CoOp, and conduct experiments in both \textit{uni} and \textit{cls} prompts types. All results are reported as {\color{cyan} the average over three runs }using three different random seeds. 

\subsection{Comparison Results}

\begin{figure}[tb]
  \centering
  \includegraphics[width=0.97\textwidth]{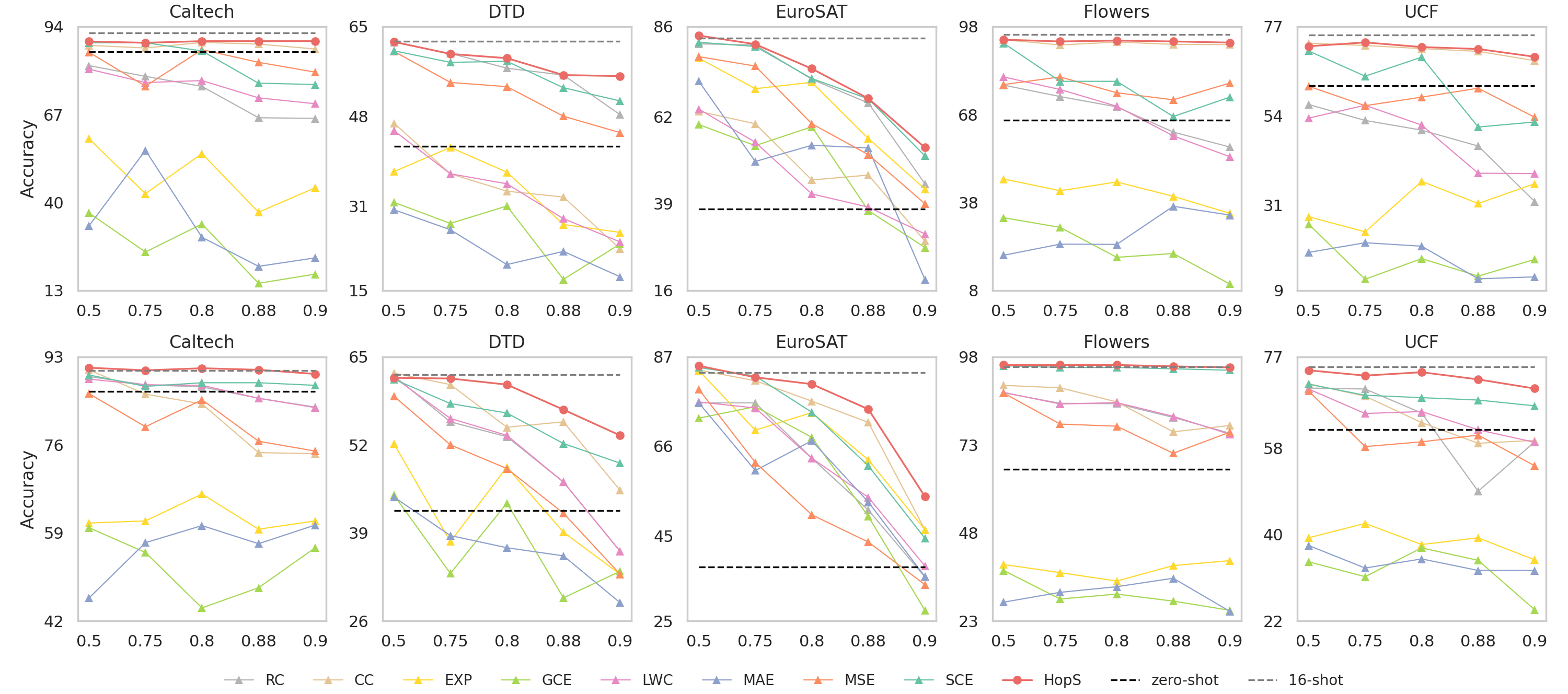}
  \caption{HopS achieves the best testing accuracy under \textit{uni}-prompt (top) and \textit{cls}-prompts (bottom) across five \textit{rand} confusion rates.}
  \label{uni_rand}
\end{figure}

\textbf{\textit{Rand}-Confusion Type.} We evaluate all methods under two prompt settings: \textit{uni}-prompt and \textit{cls}-prompts. As shown in Fig.~\ref{uni_rand}, HopS achieves SOTA performance across datasets under varying levels of label confusion. It is worth highlighting that, on Caltech, Flowers, and UCF, HopS achieves performance comparable to fully supervised 16-shot CoOp and significantly outperforms the zero-shot capability of CLIP across all datasets. Moreover, {\color{cyan} HopS maintains consistently high performance even under a severe label confusion rate of 0.9}. This highlights the effectiveness of our holistic label selection strategy in reducing label ambiguity and providing reliable supervision. The detailed numerical results can be found in Appendix Tabs.~\ref{rand-uni} and \ref{rand-cls}.

\begin{figure}[tb]
  \centering
  \includegraphics[width=\textwidth]{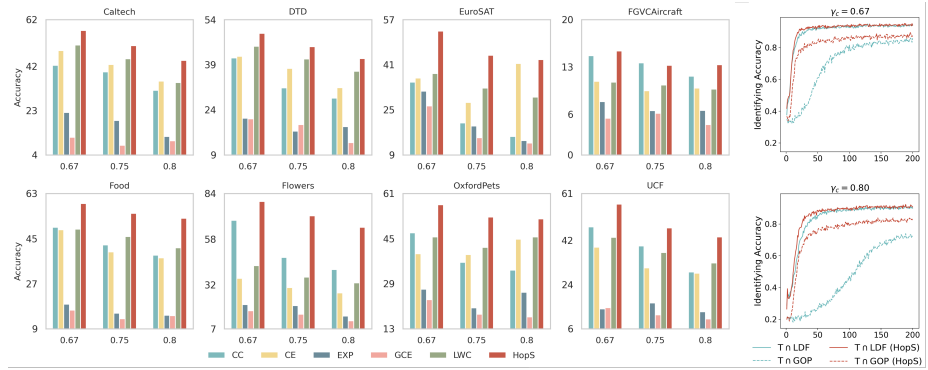} 
  \caption{HopS achieves the best averaged testing accuracy across three confusion rates (left). The guiding effect of LDF on GOP (right).}
  \label{uni_cls_insd}
\end{figure}

\noindent\textbf{\textit{Insd}-Confusion Type.} Since partial-label losses perform well under the \textit{rand} confusion setting, we further compare HopS against them under instance-dependent partial labels, which more closely resemble real-world scenarios where label noise depends on feature-level similarities. As shown in Fig.~\ref{uni_cls_insd} (left), {\color{cyan} HopS consistently outperforms other methods} under the \textit{uni}-prompt setting when averaging results over ResNet-18, ResNet-50, and CLIP-ResNet50 simulating backbones. Notably, our method retains a substantial advantage even as the confusion rate rises to 0.8. This demonstrates the robustness of HopS in more challenging and realistic conditions. 

\begin{figure}[tb]
  \centering
  \begin{minipage}[]{0.49\textwidth}
    \centering
    \includegraphics[width=\linewidth]{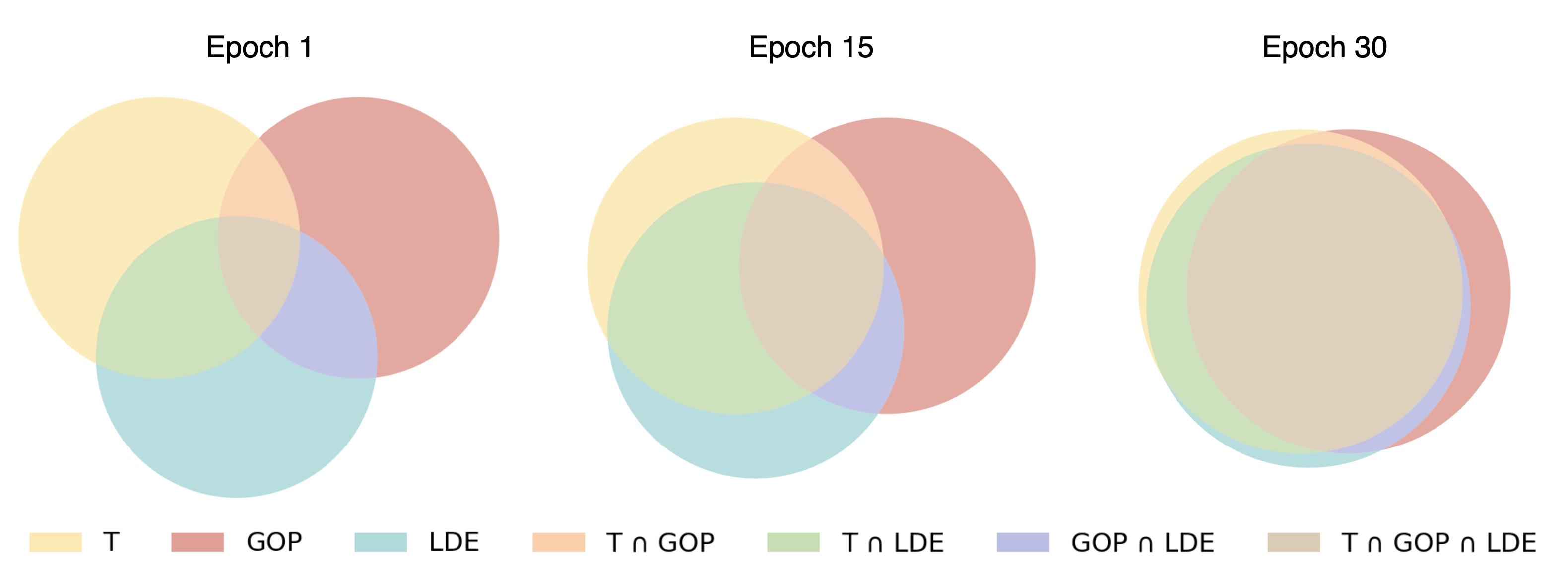}
    \caption{Overlap of labels identified by LDF and GOP in HopS with the GT \textbf{T} on the Food (\(\gamma_c = 0.67\)).}
    \label{venn}
  \end{minipage}
  \hfill
  \begin{minipage}[]{0.49\textwidth}
    \centering
    \includegraphics[width=0.95\linewidth]{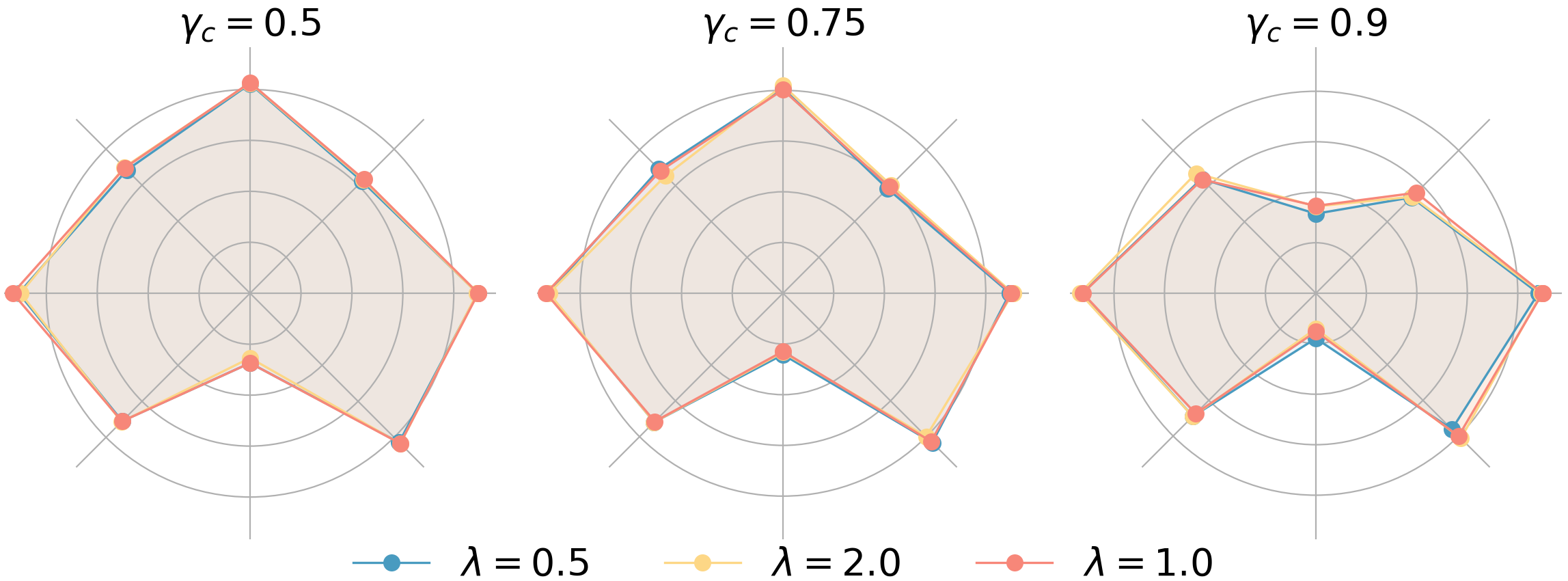} 
    \caption{The performance remains stable across a wide range of $\lambda$ on \textbf{eight} datasets (directions), and $\lambda{=}1$ works well.}
    \label{weight}
  \end{minipage}
\end{figure}
In particular, this performance gain holds across settings where candidate labels are derived from annotators with varying levels of domain expertise (i.e., simulated through different backbone encoders that reflect distinct data conditions), indicating that HopS is resilient to the type and quality of label noise. This robustness stems from the complementary nature of LDF and GOP components in HopS. These results indicate that HopS is effective not only in synthetic settings but also in real-world scenarios with instance-dependent ambiguity, such as crowd-sourced labels or low-resource domains. \textit{The detailed per-configuration results are reported in Appendix Tabs}~\ref{tab:insd_2} \textit{and}~\ref{tab:insd_1}.

\subsection{Further Analysis}
\noindent\textbf{Complement Effect of LDF and GOP.} To assess the interaction between the LDF and GOP modules, we compare their individual performance and that of the holistic model, HopS, \textbf{in identifying the ground-truth (GT) label}. The training accuracy of LDF, GOP, and HopS over 200 epochs, evaluated under the \textit{uni}-prompt on the Food dataset at confusion rates of 0.67 and 0.80, is illustrated in Fig.~\ref{uni_cls_insd} (right), emphasizing {\color{cyan} LDF's role in accelerating early optimization by guiding the global module}. The four curves compare LDF and GOP used stand-alone versus used as components within HopS. The symbol $\cap$ denotes the proportion of predicted labels from the LDF or GOP that are consistent with the GT label, where T denotes the GT label. Using modules as HopS components yields faster convergence and a higher, more stable plateau, and this advantage becomes increasingly pronounced as the $\gamma_c$ intensifies. \textit{Additional results are provided in Appendix Figs.}~\ref{fig:cal_3}-\ref{fig:ucf_3}. 

Furthermore, Fig.~\ref{venn} shows that {\color{cyan} the proportion of correct labels jointly identified by both modules (i.e., the brown region) steadily increases}, indicating a strong complementary effect. We also examine the relative contributions of the two pseudo-label sources by varying the loss weighting coefficient $\lambda \in \{0.5, 1.0, 2.0\}$. As shown in Fig.~\ref{weight}, achieving a balanced combination of local and global signals leads to consistent improvements in performance across different confusions. In contrast, placing excessive weight on either signal—whether local or global—results in performance degradation. This finding underscores the critical importance of maintaining an effective synergy between the two sources of information. \textit{The specific results are provided in Tab.}~\ref{tab:abs} \textit{in the Appendix}.

\begin{figure}[tb]
  \centering
  \begin{minipage}[]{0.43\textwidth}
    \centering
    \includegraphics[width=\linewidth]{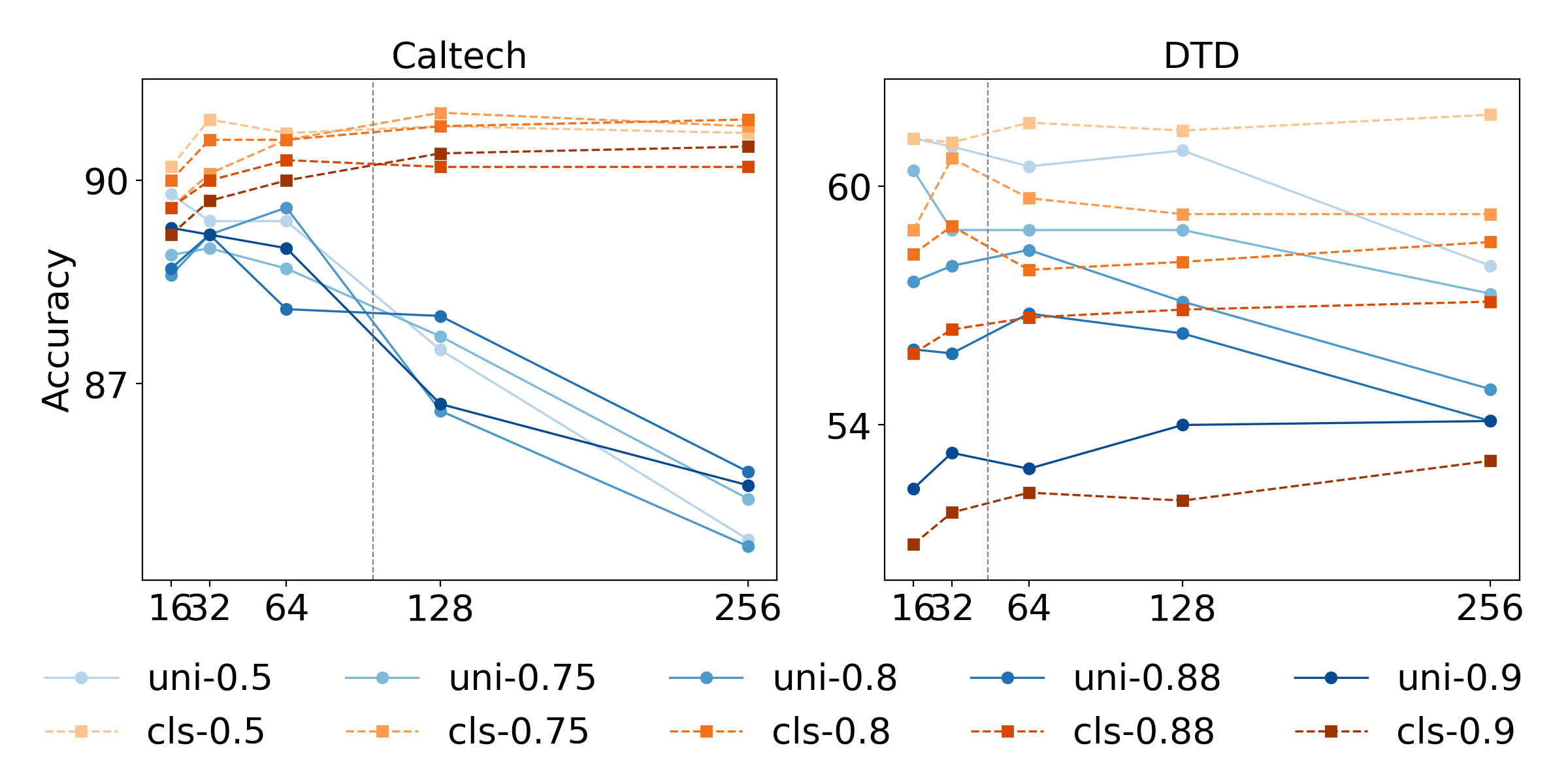}
    \captionof{figure}{$B$ close to $C$ is recommended.}
    \label{bs}
  \end{minipage}
  \hfill
  \begin{minipage}[]{0.52\textwidth}
    \centering
    \captionof{table}{Testing Accuracy (\%) of Various Candidate-Searching Strategies for label refinement in LDF.}
    \label{for-lde}
    \resizebox{\linewidth}{!}{
    \begin{tabular}{@{\hskip 5pt}c@{\hskip 5pt}||@{\hskip 5pt}c@{\hskip 5pt}||@{\hskip 5pt}c@{\hskip 5pt} c@{\hskip 5pt} c@{\hskip 5pt}|@{\hskip 5pt}c@{\hskip 5pt}}
      \toprule
      \(\mathbf{\gamma_c}\) & CoOp & Nei (Hard) & Con & Gra & Soft \\ \midrule 0.50 & 84.70 & \textbf{86.71} & 86.20 & 83.40 & 86.70 \\ 0.67 & 80.00 & 85.60 & 84.30 & 83.10 & \textbf{86.00} \\ 0.80 & 74.60 & 84.10 & 83.80 & 78.40 & \textbf{84.80} \\ \bottomrule
    \end{tabular}}
  \end{minipage}
\end{figure}

\noindent\textbf{Effect of Batch Size.} We investigate the effect of varying batch sizes, $B$, within the set $\{16, 32, 64, 128, 256\}$ for GOP, on the performance of HopS, focusing on the testing accuracy under varying levels of label confusion: 0.50, 0.75, 0.80, 0.88, and 0.90. As illustrated in Fig.~\ref{bs}, the results compare the performance of \textit{uni}-prompt and \textit{cls}-prompts across different confusion rates. For the 47-class small-scale dataset DTD and the 100-class large-scale dataset Caltech, the accuracy trends with respect to $B$ remain consistent across different confusion levels. Moreover, {\color{cyan} the most significant fluctuations in accuracy are observed when the batch size is approximately equal to the number of classes}. Based on this observation, we recommend choosing a batch size close to the number of classes. \textit{Results for other datasets are provided in the Appendix Fig.}~\ref{fig:bs} \textit{and Tab.}~\ref{tab:bs}.

\noindent\textbf{Effect of LDF.} We compare three candidate-searching strategies for label refinement in partial label learning: (1) Neighbor-based refinement (Nei), which selects pseudo-labels based on feature similarity to nearest neighbors; (2) Confidence-based refinement (Con), which chooses the most confident prediction; and (3) Graph-based propagation (Gra), which propagates labels through a constructed similarity graph, enabling global label consistency by considering relationships between all samples in the dataset and effectively transferring labels across connected nodes. As shown in Tab.~\ref{for-lde}, the simplest strategy—neighbor-based refinement—yields the best performance, suggesting that in weakly supervised few-shot settings, leveraging local structural similarity can outperform more complex propagation methods. One possible reason for this observation is that {\color{cyan} neighbor-based methods are inherently more robust to noisy predictions}. By grounding pseudo-label selection in local feature consistency, these methods can better capture semantic regularities in the data. In contrast, graph-based methods might cause noise accumulation, especially in low-data regimes, where the constructed graph may not accurately reflect the underlying label manifold. 

Additionally, we evaluate two label update strategies: Hard-KNN, which uses majority voting, and Soft-KNN, which uses similarity-weighted voting. Although Soft-KNN slightly outperforms Hard-KNN in some cases, the gains are marginal, indicating that the simple hard voting mechanism is sufficient and more computationally efficient.

\begin{figure}[htbp]
  \centering
  \begin{minipage}[t]{0.49\textwidth}
    \centering
    \includegraphics[width=0.98\linewidth]{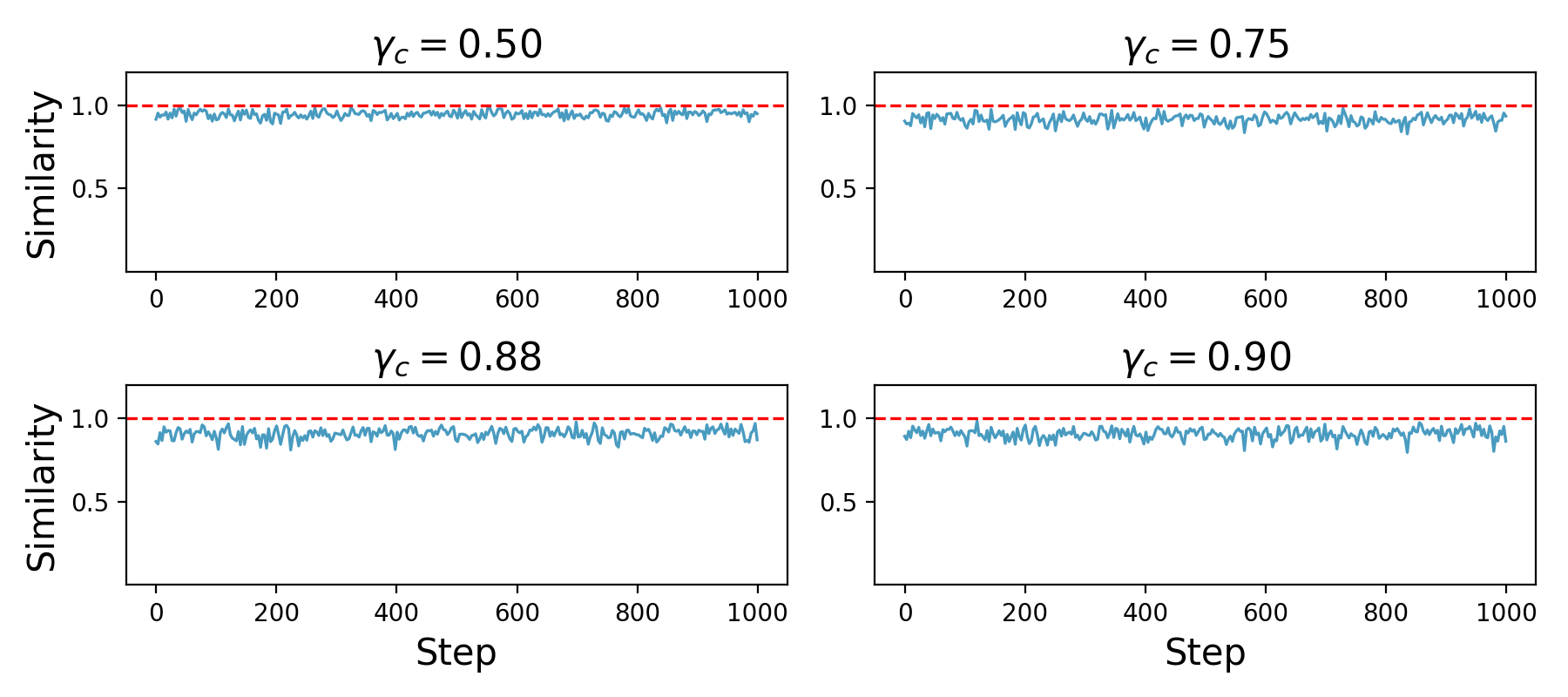}
    \captionof{figure}{High cosine similarity between candidate and ground-truth label distributions across batches.}
    \label{gop}
  \end{minipage}
  \hfill
  \begin{minipage}[t]{0.49\textwidth}
      \centering
      \includegraphics[width=0.90\linewidth]{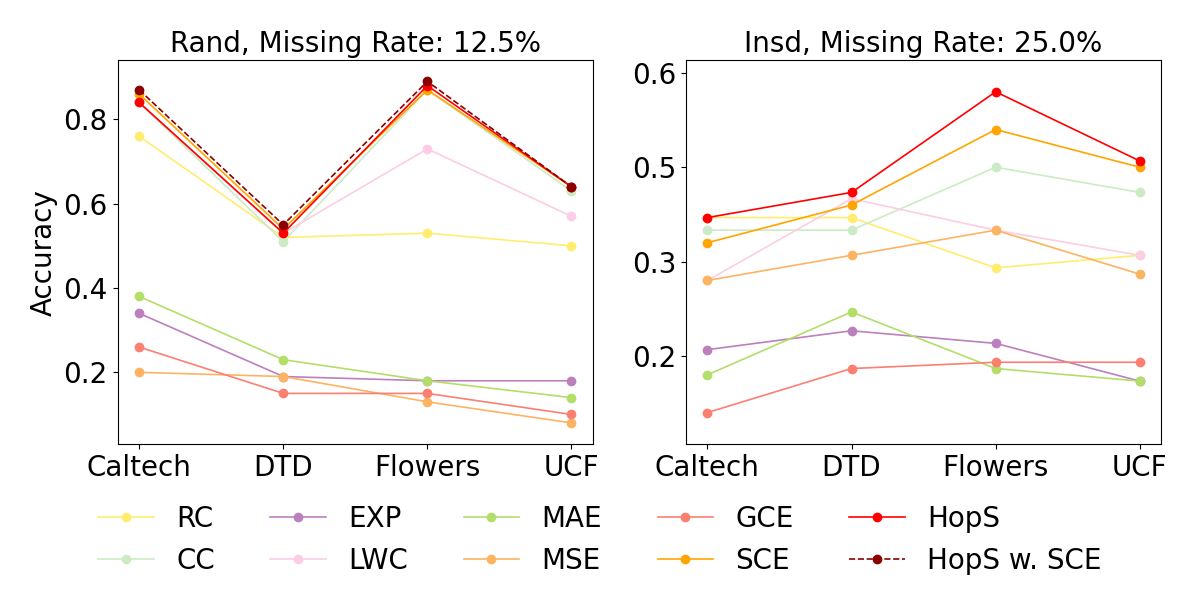} 
      \caption{Testing accuracy under the \textit{uni}-prompt across two noisy conditions with different confusion rates.}
      \label{noise}
  \end{minipage}
\end{figure}

\noindent\textbf{Effect of GOP.} The GOP module assumes that the candidate label distribution closely resembles the ground-truth label distribution. To assess this consistency, we use cosine similarity, which captures the directional alignment between two distributions and reflects their structural similarity regardless of scale. This metric provides a reliable measure of how well the candidate labels approximate the true semantic distribution. As shown in Fig.~\ref{gop}, {\color{cyan} the cosine similarity approaches 1 on EuroSAT under four different confusion rates, indicating a strong alignment between the distributions}. We opt not to use Kullback–Leibler (KL) divergence due to the prevalence of zero values in the ground-truth distributions, which makes KL computation unstable and ill-defined. In contrast, cosine similarity is robust to sparsity.


\noindent\textbf{Comparison with Full-Data PLL Methods.} 
We conducted a systematic comparison with two recently proposed partial label learning methods, Papi\cite{xia2023towards} and CroSel\cite{tian2024crosel}. Unlike the 16-shot setting used in previous experiments, Papi and CroSel were trained on the entire datasets to fully exploit their optimal performance, while HopS was still trained under the 16-shot few-shot setting. The experiments were conducted on two challenging datasets, Caltech and Flowers, both characterized by a large number of categories, with Flowers being more fine-grained and exhibiting higher intra-class similarity. As shown in Tab.~\ref{tab:confusion_compare}, the left and right sub-columns for each dataset correspond to \textit{rand} and \textit{insd}, respectively. {\color{cyan} HopS’s strong performance across both datasets and three confusion rates}, demonstrating its robustness in handling both diverse and fine-grained categories. Due to space limitations, we report only the results with the ResNet-50 backbone; results for other backbones are provided in Appendix Tab.~\ref{tab:threebackbones}.


\begin{table}[t]
  \centering
  \caption{HopS achieves the best testing accuracy (\%) under two confusion types.}
  \label{tab:confusion_compare}
  \resizebox{0.9\linewidth}{!}{
  \begin{tabular}{@{\hskip 5pt}c@{\hskip 5pt}||ccc|ccc||ccc|ccc}
    \toprule
    \multirow{3}{*}{\(\mathbf{\gamma_c}\)} 
      & \multicolumn{6}{c||}{Caltech} 
      & \multicolumn{6}{c}{Flowers} \\
    \cmidrule(lr){2-7}\cmidrule(l){8-13}
    & HopS & CroSel & Papi & HopS & CroSel & Papi
    & HopS & CroSel & Papi & HopS & CroSel & Papi \\
    \midrule
    0.67 & \textbf{89.0} & 84.0 & 55.9 & \textbf{50.3} & 39.0 & 42.0
         & \textbf{92.9} & 92.1 & 78.7 & \textbf{82.0} & 78.0 & 67.0 \\
    0.75 & \textbf{89.0} & 82.7 & 51.8 & \textbf{45.3} & 25.2 & 23.5
         & \textbf{92.9} & 90.5 & 70.4 & \textbf{74.0} & 62.0 & 58.7 \\
    0.80 & \textbf{89.5} & 73.9 & 52.2 & \textbf{38.9} & 11.8 & 13.4
         & \textbf{93.2} & 89.0 & 57.8 & \textbf{65.0} & 50.2 & 49.8 \\
    \bottomrule
  \end{tabular}
  }
\end{table}

\noindent\textbf{Settings with Missing Ground-Truth.} 
To gain deeper insights into the robustness of HopS, we evaluate it under more challenging settings where the ground-truth label may be absent from the candidate set, i.e., the noisy partial label learning scenario~\cite{xu2023alim}. We conduct 16-shot experiments on two types of label confusion, \textit{rand} and \textit{insd}, as previously described, with each candidate set $S$ containing three labels. The missing rates of ground-truth labels are 12.5\% (2 out of 16 shots) for \textit{rand} and 25\% (4 out of 16 shots) for \textit{insd}. 

As shown in Fig.~\ref{noise}, under the \textit{insd} setting, HopS consistently uncovers meaningful correlations within the candidate label sets, leading to significantly better performance than other methods. This suggests that {\color{cyan} HopS is particularly well-suited for real-world scenarios where label noise in SS is instance-dependent}. In contrast, under the \textit{rand} setting where such dependencies are absent, HopS performs slightly worse than the SOTA robust loss. Nevertheless, its performance can be improved by integrating robust loss functions (e.g., the SCE loss). Figure~\ref{noise} (left) illustrates that HopS, when combined with the SCE loss, achieves the best performance under the \textit{rand} setting, highlighting its flexibility and compatibility with other robust learning techniques. \textit{Detailed values corresponding to Fig.}~\ref{noise} \textit{can be found in Appendix Tabs.}~\ref{tab:4noise} \textit{and} \ref{tab:2noise}.

\noindent\textbf{Challenging Settings with Long-tail Distribution.} To further evaluate robustness under distribution shift, we compare HopS with SoLar \cite{wang2022solar} on two representative datasets under two long-tailed patterns: exponentially decayed class-frequency distribution ($^e$) and two-level distribution ($^s$). As summarized in Table~\ref{tab:lt_set}, SoLar suffers a significant performance drop in the data-limited few-shot setting, as it relies on learning sufficiently strong representations. In contrast, {\color{cyan}HopS consistently outperforms SoLar across all long-tailed configurations}, regardless of whether uni-prompts (HopS$^u$) or cls-prompts (HopS$^c$) are used.



\begin{figure*}[tb]
  \centering
  \begin{minipage}[t]{0.75\textwidth}
    \centering
    \captionof{table}{Testing accuracy (\%) under under long-tailed partial-label settings with different confusion types.}
    \label{tab:lt_set}
    \resizebox{\linewidth}{!}{
    \begin{tabular}{@{\hskip 3pt}c||c||cccc||cccc@{\hskip 3pt}}
      \toprule
      \multirow{2}{*}{Dataset} & \multirow{2}{*}{Method}
        & \multicolumn{4}{c||}{$\gamma_c=0.67$} & \multicolumn{4}{c@{}}{$\gamma_c=0.75$} \\
      \cmidrule(lr){3-6}\cmidrule(lr){7-10}
        & & \textit{insd}$^{e}$ & \textit{insd}$^{s}$ & \textit{rand}$^{e}$ & \textit{rand}$^{s}$
          & \textit{insd}$^{e}$ & \textit{insd}$^{s}$ & \textit{rand}$^{e}$ & \textit{rand}$^{s}$ \\
      \midrule
      \multirow{3}{*}{Caltech}
        & SoLar      & 9.2  & 7.9  & 16.1 & 22.6 & 6.4  & 4.4  & 18.5 & 19.4 \\
        & HopS$^{u}$ & 46.4 & 50.7 & 71.7 & 71.2 & 35.6 & 42.0 & 67.6 & 67.4 \\
        & HopS$^{c}$ & 50.9 & 52.8 & 74.6 & 71.4 & 42.4 & 45.9 & 72.6 & 70.7 \\
      \midrule
      \multirow{3}{*}{Flowers}
        & SoLar      & 11.1 & 14.0 & 18.1 & 20.8 & 9.1  & 10.6 & 17.3 & 20.7 \\
        & HopS$^{u}$ & 48.3 & 51.4 & 65.4 & 61.1 & 39.8 & 43.8 & 61.9 & 53.9 \\
        & HopS$^{c}$ & 54.4 & 53.1 & 69.9 & 66.1 & 46.8 & 47.5 & 69.9 & 59.7 \\
      \bottomrule
    \end{tabular}}
  \end{minipage}
  \hfill
  \begin{minipage}[t]{0.225\textwidth}
    \centering
    \captionof{table}{Comparison of epoch time.}
    \label{tab:time_comparison}
    \resizebox{\linewidth}{!}{
    \begin{tabular}{@{\hskip 3pt}c@{\hskip 3pt}|@{\hskip 3pt}c@{\hskip 3pt}}
      \toprule
      Method & Time (s) \\
      \midrule
      EXP  & 0.12 \\
      MSE  & 0.11 \\
      GCE  & 0.11 \\
      LWC  & 0.11 \\
      MAE  & 0.12 \\
      SCE  & 0.11 \\
      \midrule
      HopS & 0.12 \\
      \bottomrule
    \end{tabular}}
  \end{minipage}
\end{figure*}

\noindent\textbf{Running Time of Methods.} We measured the training time computational cost of each method, reported as the average time per epoch (in seconds), as shown in Tab.~\ref{tab:time_comparison}. Notably, although HopS introduces both a local module and a global module to enhance label identification, {\color{cyan}the additional computational overhead remains negligible due to the relatively small dataset size}. As a result, HopS achieves superior performance without incurring a significant increase in training cost. \textit{Additional results are provided in Tab.}~\ref{tab:training_time}\textit{ of Appendix}.
\section{Conclusion}
\label{sec:conclusion}
In this work, we propose a holistic label selection framework for prompt learning under partial-label supervision, which improves robustness by combining a local density-based filter with a global optimal transport planner. By leveraging the generalization ability of frozen pre-trained vision-language encoders, HopS enables effective label disambiguation. Extensive experiments on eight benchmark datasets demonstrate consistent and significant improvements over strong baselines, highlighting its practicality and generalizability under weak supervision. 

\section*{Acknowledgements}
The work is supported in part by Shandong Sci-tech SMEs Innovation Project
(No.\;2024TSGC0740), Natural Science Foundation of China (No.\;U23A20389), Natural Science Foundation of Shandong Province (No.\;ZR2024MF101), and Young Expert of Taishan Scholars (No.\;tsqn202312026).
\bibliographystyle{splncs04}
\bibliography{main}

\newpage
\appendix

\section{Experimental Settings}
\label{sec:exp_set}
\subsection{Confusion Types.} To simulate realistic partial label learning scenarios, we assign each training instance a candidate label set consisting of the ground-truth label and \(L\)-1 confusing labels, thereby introducing controlled ambiguity. \textbf{1}) Generation of \textit{rand}-confusion, randomly samples \(L\)-1 incorrect labels for each instance. To promote diversity, previously used confusing labels for the same class are avoided when possible. If the number of available incorrect labels is insufficient, sampling with replacement is applied. \textbf{2}) Generation of \textit{insd}-confusion, intentional selects confusion labels based on visual similarity. Specifically, image features are first extracted using a pre-trained model, and a prototype vector is computed for each class. For each instance, the cosine similarity between its feature representation and all class prototypes is calculated. Then, the top-(\(L\)-1) most similar labels—excluding the ground-truth—are selected, contributing to more challenging and informative candidate label sets with visually similar confusing labels. 


\subsection{Implementation Details.} A \textbf{16-shot} training set is randomly sampled from each dataset in a class-balanced manner, with a fixed random seed to ensure experimental reproducibility. The original test set is used for evaluation. Model training is conducted using the \textbf{SGD optimizer} with a \textbf{learning rate} of \textbf{0.002 for a maximum of 200 epochs}. A cosine annealing schedule is applied to the learning rate, with a constant warm-up set to \textbf{1e-5 for the first epoch}. The \textbf{batch size} is \textbf{32}, and the training objective is the standard cross-entropy loss. The model adopts ResNet-50 as the backbone, configured with random initialization, \textbf{16 context tokens}, and the end token position for classification. In terms of module-specific hyperparameters, the \textbf{confidence threshold \(\tau\) = 0.4} and the number of \textbf{nearest neighbors \(k\) = 20} are used in the LDF module. For the GOP module, the \textbf{cost matrix} in the Sinkhorn algorithm is scaled by an \textbf{entropy regularization coefficient \(\varepsilon\) = 0.05}, and the number of \textbf{Sinkhorn iterations} is set to \textbf{50}. The \textbf{weighting coefficient} for the two loss components is set to \textbf{\(\lambda\) = 1.0}. Regarding computational resources, all experiments are conducted on a Linux-based system equipped with eight NVIDIA RTX 4090 GPUs, with each model requiring approximately 18–30 minutes for training and peaking at 6696 MiB of memory usage. The software environment includes Python 3.8, PyTorch 1.13.1, and CUDA 12.0, along with essential libraries such as NumPy 1.21.6, Pandas 1.3.5, scikit-learn 1.0.2, and torchvision 0.14.1.

\subsubsection{Dataset.} We adopt eight datasets to comprehensively evaluate our method. Caltech and EuroSAT are general classification datasets covering diverse object and scene categories. In contrast, DTD, FGVCAircraft, Food, Flowers, OxfordPets, and UCF focus on fine-grained classification, emphasizing subtle differences within textures, aircraft models, food dishes, flower species, pet breeds, and human actions respectively. Table~\ref{ori-data} presents the original performance across eight datasets, evaluated using two models: the zero-shot CLIP model with hand-crafted prompts ( 0-clip\(^h\)), which requires no training, and the 16-shot CoOp model with learnable prompts trained on precise labels. The learnable prompts in the 16-shot CoOp model are further divided into two variants: unified prompt (\textit{uni}), which use a shared context vector across all classes ( 16-coop\(^u\)), and classified prompts (\textit{cls}), which assign distinct context vectors to each class ( 16-coop\(^c\)). All results are reproduced by ourselves to ensure a consistent comparison.

\begin{table}
\centering
\caption{Testing accuracy (\%) of models under the \textit{uni}-prompt and \textit{cls}-prompts.}
\label{ori-data}
\resizebox{\linewidth}{!}{
\begin{tabular}{@{\hskip 5pt}c||c||c||c|c|c@{\hskip 5pt}}
\toprule
Dataset      & Class & Hand-crafted prompt                             & 0-clip\(^{h}\) & 16-coop\(^{u}\) & 16-coop\(^{c}\) \\ \midrule
EuroSAT      & 10    & a centered satellite photo of {[}class{]}     & 37.6                        & 82.9                       & 83.3                       \\
OxfordPets   & 37    & a photo of a {[}class{]}, a type of pet.      & 85.7                        & 86.5                       & 78.9                       \\
DTD          & 47    & {[}class{]} texture.                          & 42.3                        & 62.2                       & 62.4                       \\
Caltech   & 100   & a photo of a {[}class{]}.                     & 86.3                        & 92.0                       & 90.4                       \\
FGVCAircraft & 100   & a photo of a {[}class{]}, a type of aircraft. & 17.0                        & 31.8                       & 37.7                       \\
Food      & 101   & a photo of a {[}class{]}, a type of food.     & 77.4                        & 74.5                       & 70.4                       \\
UCF       & 101   & a photo of a person doing {[}class{]}.        & 61.8                        & 74.8                       & 74.9                       \\
Flowers   & 102   & a photo of a {[}class{]}, a type of flower.   & 66.1                        & 95.2                       & 95.1                       \\ \bottomrule
\end{tabular}}
\end{table}

\subsubsection{Baselines.} 
\label{sec:exp_set_basl}
We compare HopS with eight state-of-the-art loss functions from partial-label learning and robust learning. All hyperparameters for these methods are carefully tuned according to the settings reported in their original publications, and the configurations of all methods are detailed in Table~\ref{tab:methods}. 

In addition, we compare HopS with three recent PLL methods, CroSel, PaPi, and SoLar, each of which adopts a different strategy for handling label ambiguity. CroSel utilizes cross-entropy loss to train the label selection process, while incorporating consistency regularization (with data augmentation and MixUp) to reduce selection noise. PaPi combines a prototypical alignment loss (based on KL divergence) with cross-entropy loss to optimize class representations and improve ambiguity handling. SoLar is a long-tailed partial-label learning method that improves robustness under class imbalance by leveraging image representation optimization and label-distribution refinement.

\begin{sidewaystable}[]
\centering
\caption{Partial label learning losses and robust learning losses}
\label{tab:methods}
\resizebox{0.9\linewidth}{!}{
\begin{tabular}{@{\hspace{5pt}} c|| c|| >{\centering\arraybackslash}p{10cm} @{\hspace{5pt}}}
\toprule
Method & Loss formula & Description \\[2pt] \midrule

RC & 
\(\displaystyle \mathcal{L}_{\text{RC}} = -\frac{1}{B} \sum_{i=1}^B \sum_{j=1}^C \log(p_{ij}) \cdot y_{ij}\) 
& Risk-consistent PLL method that assumes the candidate label set is uniformly sampled from the true label. \\[15pt] \midrule

CC & 
\(\displaystyle \mathcal{L}_{\text{CC}} = -\frac{1}{B} \sum_{i=1}^B \log\left( \sum_{j=1}^C p_{ij} \cdot y_{ij}\right)\) 
& Classifier-consistent PLL method under the same assumption as RC. \\[15pt] \midrule

EXP & 
\(\displaystyle \mathcal{L}_{\text{EXP}} = \frac{1}{B} \sum_{i=1}^B \exp\left( -\sum_{j=1}^C p_{ij} \cdot y_{ij} \right)\)
& PLL method that adopts the exponential loss as a risk estimator. \\[15pt] \midrule

GCE & 
\(\displaystyle \mathcal{L}_{\text{GCE}} = \frac{1}{B} \sum_{i=1}^B \sum_{j=1}^C (1 - p_{ij}) \cdot y_{ij}\) 
& Robust loss generalizing cross-entropy and mean absolute error to improve tolerance to label noise. \\[15pt] \midrule

LWC & 
\(\displaystyle
\begin{aligned}
\mathcal{L}_{\text{LWC}} = - \sum_{i,j} y_{ij} \cdot \log(p_{ij}) - \frac{1}{C} \sum_{i,j} (1 - y_{ij}) \cdot \log(1 - p_{ij})
\end{aligned}
\)
& PLL method that considers trade-off between losses on candidate and non-candidate label sets. \\[15pt] \midrule

MAE & 
\(\displaystyle \mathcal{L}_{\text{MAE}} = \frac{1}{B} \sum_{i=1}^B \sum_{j=1}^C | p_{ij} - y_{ij} |\) 
& Robust loss treating all classes equally by minimizing absolute error. \\[15pt] \midrule

MSE & 
\(\displaystyle \mathcal{L}_{\text{MSE}} = \frac{1}{B} \sum_{i=1}^B \sum_{j=1}^C (p_{ij} - y_{ij})^2\) 
& PLL method that estimates risk via mean squared error between predictions and candidate labels. \\[15pt] \midrule

SCE & 
\(\displaystyle
\begin{aligned} \mathcal{L}_{\text{SCE}} = \alpha \cdot \underbrace{- \sum_{j} y_{ij} \cdot \log(p_{ij})}_{\text{Cross Entropy}} + \beta \cdot \underbrace{- \sum_{j} p_{ij} \cdot \log(y_{ij})}_{\text{Reverse CE}} \end{aligned}
\) 
& Symmetric loss combining cross-entropy and reverse cross-entropy to enhance robustness.  (\(\alpha = 0.1\), \(\beta = 1.0\))\\[15pt] \midrule

HopS(ours) & 
\(\displaystyle
\begin{aligned} \mathcal{L}_{\text{HopS}} = \underbrace{- \sum_{j} y_{ij} \cdot \log(p^{\text{LDF}}_{ij})}_{\text{LDF Cross Entropy}} + \lambda \cdot \underbrace{- \sum_{j} y_{ij} \cdot \log(p^{\text{GOP}}_{ij})}_{\text{GOP Cross Entropy}} \end{aligned}
\)
& Composite loss jointly optimizing cross-entropy from local density filtering and global optimal planning. (\(\lambda = 1.0\)) \\[15pt] \bottomrule
\end{tabular}}
\end{sidewaystable}

\section{Comparison Results}\label{comp_results}
\subsection{\textit{Rand}-Confusion Labels.} All methods are evaluated under two prompt types: the (\textit{uni}) prompt and the (\textit{cls}) prompts. The corresponding results are summarized in Tables~\ref{rand-uni} and \ref{rand-cls}, where all values are reported as percentages (the percent sign is omitted). The best performance is highlighted in \textbf{bold}, and the second-best is \underline{underlined}.


\begin{table}[htbp]
\centering
\caption{Testing accuracy (\%) under \textit{uni}-prompt across five \textit{rand} confusion rates.}
\label{rand-uni}
\begin{tabular}{@{\hskip 3pt}c||c||c|c|c|c|c|c|c|c|c@{\hskip 3pt}}
\toprule
Dataset                       & \(\gamma_c\) & RC   & CC            & EXP  & GCE  & LWC  & MAE  & MSE  & SCE           & HopS          \\ \midrule
\multirow{5}{*}{Caltech}      & 0.50  & 82.0 & 88.2          & 59.6 & 36.8 & 80.9 & 32.8 & 86.1 & {\ul 89.0}    & \textbf{89.4} \\
                              & 0.75  & 78.7 & {\ul 87.4}          & 42.7 & 24.8 & 76.8 & 56.0 & 75.7 & \textbf{89.0} & \textbf{89.0} \\
                              & 0.80  & 75.6 & {\ul 89.1}    & 55.0 & 33.4 & 77.4 & 29.3 & 86.9 & 86.5          & \textbf{89.5} \\
                              & 0.88  & 66.0 & {\ul 88.6}    & 37.0 & 15.2 & 72.1 & 20.4 & 83.0 & 76.6          & \textbf{89.5} \\
                              & 0.90  & 65.8 & {\ul 87.1}    & 44.5 & 18.0 & 70.3 & 23.0 & 80.0 & 76.2          & \textbf{89.5} \\ \midrule
\multirow{5}{*}{DTD}          & 0.50  & 46.7 & 62.0          & 37.5 & 31.7 & 45.3 & 30.3 & 60.3 & 60.4          & \textbf{62.1} \\
                              & 0.75  & 37.1 & \textbf{59.9} & 42.1 & 27.7 & 37.1 & 26.5 & 54.4 & 58.2          & {\ul 59.8}    \\
                              & 0.80  & 33.8 & 57.1          & 37.4 & 31.0 & 35.2 & 19.9 & 53.6 & {\ul 58.4}    & \textbf{59.0} \\
                              & 0.88  & 32.7 & \textbf{55.8} & 27.5 & 17.1 & 28.6 & 22.4 & 48.0 & 53.4          & \textbf{55.8} \\
                              & 0.90  & 22.9 & 48.3          & 26.0 & 23.8 & 24.2 & 17.6 & 44.9 & {\ul 50.9}    & \textbf{55.6} \\ \midrule
\multirow{5}{*}{EuroSAT}      & 0.50  & 63.5 & 81.4          & 77.6 & 60.0 & 64.1 & 71.5 & 78.0 & {\ul 81.8}    & \textbf{83.6} \\
                              & 0.75  & 60.2 & {\ul 81.1}    & 69.5 & 54.4 & 55.3 & 50.2 & 75.5 & 80.7          & \textbf{81.3} \\
                              & 0.80  & 45.3 & 72.0          & 71.2 & 59.4 & 41.6 & 54.5 & 60.2 & {\ul 72.2}    & \textbf{74.8} \\
                              & 0.88  & 46.6 & 65.7          & 56.3 & 37.2 & 38.1 & 53.8 & 52.0 & {\ul 66.8}    & \textbf{66.9} \\
                              & 0.90  & 29.1 & 44.3          & 42.9 & 27.4 & 31.0 & 18.9 & 39.0 & {\ul 51.7}    & \textbf{53.9} \\ \midrule
\multirow{5}{*}{Food}         & 0.50  & 54.2 & {\ul 67.8}    & 24.8 & 18.5 & 51.9 & 23.2 & 43.8 & 65.9          & \textbf{68.4} \\
                              & 0.75  & 46.3 & {\ul 68.0}    & 22.8 & 28.9 & 45.2 & 20.3 & 48.6 & 63.8          & \textbf{69.0} \\
                              & 0.80  & 42.3 & {\ul 65.3}    & 22.2 & 19.8 & 41.9 & 23.0 & 55.6 & 52.8          & \textbf{68.2} \\
                              & 0.88  & 33.3 & {\ul 64.4}    & 19.9 & 21.3 & 32.2 & 11.1 & 44.2 & 38.9          & \textbf{66.0} \\
                              & 0.90  & 31.3 & {\ul 62.3}    & 20.9 & 13.1 & 27.6 & 16.7 & 49.1 & 30.8          & \textbf{65.0} \\ \midrule
\multirow{5}{*}{Flowers}      & 0.50  & 78.0 & {\ul 93.5}    & 46.0 & 32.8 & 80.8 & 20.0 & 78.2 & 92.3          & \textbf{93.5} \\
                              & 0.75  & 74.1 & {\ul 91.7}    & 42.0 & 29.5 & 76.5 & 23.8 & 80.8 & 79.3          & \textbf{92.9} \\
                              & 0.80  & 70.6 & {\ul 92.7}    & 45.0 & 19.3 & 70.8 & 23.7 & 75.4 & 79.3          & \textbf{93.2} \\
                              & 0.88  & 62.0 & {\ul 91.9}    & 40.1 & 20.6 & 60.7 & 36.7 & 73.0 & 67.3          & \textbf{92.9} \\
                              & 0.90  & 56.9 & {\ul 91.8}    & 34.3 & 10.2 & 53.6 & 33.7 & 78.7 & 73.9          & \textbf{92.5} \\ \midrule
\multirow{5}{*}{UCF}          & 0.50  & 56.9 & \textbf{72.7} & 28.0 & 26.1 & 53.4 & 18.8 & 61.6 & 70.7          & {\ul 71.9}    \\
                              & 0.75  & 52.8 & {\ul 72.1}    & 24.1 & 11.9 & 56.7 & 21.3 & 56.6 & 64.2          & \textbf{72.9} \\
                              & 0.80  & 50.3 & {\ul 71.3}    & 37.1 & 17.2 & 51.6 & 20.4 & 58.8 & 69.1          & \textbf{71.7} \\
                              & 0.88  & 46.2 & {\ul 70.6}    & 31.4 & 12.7 & 39.2 & 12.0 & 61.1 & 51.1          & \textbf{71.2} \\
                              & 0.90  & 31.9 & {\ul 68.2}    & 36.5 & 17.0 & 39.1 & 12.5 & 53.6 & 52.4          & \textbf{69.2} \\ \midrule 
\multirow{5}{*}{FGVCAircraft} & 0.50  & 19.5 & \textbf{27.3} & 9.5  & 6.8  & 22.7 & 12.0 & 18.3 & 25.0          & {\ul 26.8}    \\
                              & 0.75  & 17.0 & \textbf{25.5} & 8.3  & 6.3  & 17.7 & 5.2  & 19.4 & 18.5          & {\ul 24.6}    \\
                              & 0.80  & 15.4 & \textbf{24.7} & 11.2 & 6.2  & 14.3 & 5.4  & 19.4 & 17.6          & {\ul 23.8}    \\
                              & 0.88  & 11.2 & \textbf{20.4} & 10.7 & 5.0  & 11.4 & 4.4  & 16.8 & 13.0          & {\ul 20.2}    \\
                              & 0.90  & 11.4 & \textbf{20.4} & 10.9 & 4.1  & 10.0 & 7.4  & 14.6 & 10.6          & {\ul 19.3}    \\ \midrule
\multirow{5}{*}{OxfordPets}   & 0.50  & 58.9 & 82.5          & 52.3 & 38.0 & 60.6 & 14.7 & 82.0 & \textbf{84.0} & {\ul 83.5}    \\
                              & 0.75  & 49.4 & {\ul 83.0}    & 46.9 & 31.0 & 49.4 & 46.6 & 76.2 & 82.6          & \textbf{83.8} \\
                              & 0.80  & 42.4 & 80.2          & 49.3 & 22.7 & 40.9 & 26.3 & 72.5 & {\ul 80.9}    & \textbf{82.6} \\
                              & 0.88  & 34.8 & 73.9          & 23.7 & 39.1 & 31.3 & 38.1 & 74.2 & {\ul 76.8}    & \textbf{79.9} \\
                              & 0.90  & 22.0 & {\ul 74.2}    & 40.7 & 12.3 & 21.9 & 18.3 & 45.0 & 66.6          & \textbf{80.5} \\ \bottomrule
\end{tabular}
\end{table}

\begin{table}[htbp]
\centering
\caption{Testing accuracy (\%) under \textit{cls}-prompts across five \textit{rand} confusion rates.}
\label{rand-cls}
\begin{tabular}{@{\hskip 3pt}c||c||c|c|c|c|c|c|c|c|c@{\hskip 3pt}}
\toprule
Dataset                       & \(\gamma_c\) & RC            & CC            & EXP  & GCE  & LWC        & MAE  & MSE  & SCE           & HopS          \\ \midrule
\multirow{5}{*}{Caltech}      & 0.50                  & 89.2          & {\ul 90.5}    & 60.9 & 60.0 & 88.7       & 46.4 & 85.9 & 89.5          & \textbf{90.9} \\
                              & 0.75                  & 87.6          & 85.8          & 61.3 & 55.2 & {\ul 87.6} & 57.1 & 79.5 & 87.3          & \textbf{90.4} \\
                              & 0.80                  & 87.2          & 83.9          & 66.5 & 44.5 & 87.5       & 60.4 & 84.7 & {\ul 88.0}    & \textbf{90.8} \\
                              & 0.88                  & 85.0          & 74.5          & 59.7 & 48.4 & 85.0       & 56.9 & 76.7 & {\ul 88.0}    & \textbf{90.5} \\
                              & 0.90                  & 83.3          & 74.3          & 61.3 & 56.1 & 83.2       & 60.5 & 74.8 & {\ul 87.5}    & \textbf{89.7} \\ \midrule
\multirow{5}{*}{DTD}          & 0.50                  & {\ul 62.2}    & \textbf{62.6} & 52.2 & 44.6 & 62.0       & 44.3 & 59.2 & 61.7          & 61.9          \\
                              & 0.75                  & 55.4          & {\ul 60.9}    & 37.7 & 33.0 & 55.9       & 38.6 & 52.0 & 58.1          & \textbf{61.8} \\
                              & 0.80                  & 53.2          & 54.6          & 48.7 & 43.4 & 53.4       & 36.8 & 48.5 & {\ul 56.7}    & \textbf{60.9} \\
                              & 0.88                  & 46.5          & {\ul 55.4}    & 39.1 & 29.4 & 46.5       & 35.6 & 41.9 & 52.2          & \textbf{57.2} \\
                              & 0.90                  & 36.3          & 45.3          & 32.9 & 33.3 & 36.3       & 28.7 & 32.9 & {\ul 49.3}    & \textbf{53.4} \\ \midrule
\multirow{5}{*}{EuroSAT}      & 0.50                  & 76.2          & 83.9          & 83.8 & 72.6 & 76.4       & 76.2 & 79.4 & 84.5          & \textbf{84.9} \\
                              & 0.75                  & 76.2          & 81.4          & 69.8 & 75.3 & 75.0       & 60.3 & 62.2 & \textbf{82.5} & {\ul 82.2}    \\
                              & 0.80                  & 63.2          & {\ul 76.6}    & 73.9 & 68.1 & 63.2       & 67.3 & 49.9 & 74.0          & \textbf{80.6} \\
                              & 0.88                  & 51.0          & {\ul 71.6}    & 62.8 & 49.5 & 54.0       & 52.9 & 43.5 & 61.4          & \textbf{74.7} \\
                              & 0.90                  & 35.3          & {\ul 46.4}    & 46.4 & 27.5 & 37.9       & 35.4 & 33.5 & 44.4          & \textbf{54.2} \\ \midrule
\multirow{5}{*}{Food}         & 0.50                  & 66.9          & 68.6          & 33.7 & 31.4 & 66.5       & 34.7 & 63.5 & {\ul 68.7}    & \textbf{68.8} \\
                              & 0.75                  & 60.9          & 62.9          & 33.5 & 27.5 & 61.2       & 26.8 & 58.0 & {\ul 66.4}    & \textbf{69.0} \\
                              & 0.80                  & 59.2          & 57.2          & 38.8 & 24.7 & 59.4       & 28.8 & 58.9 & {\ul 66.1}    & \textbf{68.4} \\
                              & 0.88                  & 54.2          & 57.2          & 34.0 & 27.7 & 54.4       & 25.0 & 49.7 & {\ul 65.0}    & \textbf{68.1} \\
                              & 0.90                  & 49.0          & 52.6          & 30.8 & 24.1 & 48.7       & 24.0 & 48.4 & {\ul 63.8}    & \textbf{67.1} \\ \midrule
\multirow{5}{*}{Flowers}      & 0.50                  & 87.8          & 89.9          & 39.0 & 37.3 & 87.9       & 28.3 & 87.7 & {\ul 95.5}    & \textbf{95.7} \\
                              & 0.75                  & 84.8          & 89.2          & 36.7 & 29.2 & 84.6       & 31.1 & 78.9 & {\ul 95.0}    & \textbf{95.7} \\
                              & 0.80                  & 84.7          & 85.2          & 34.3 & 30.6 & 85.0       & 32.7 & 78.3 & {\ul 95.0}    & \textbf{95.7} \\
                              & 0.88                  & 80.8          & 76.7          & 38.7 & 28.6 & 81.1       & 35.1 & 70.6 & {\ul 94.6}    & \textbf{95.3} \\
                              & 0.90                  & 76.3          & 78.5          & 40.1 & 26.0 & 76.0       & 25.7 & 76.6 & {\ul 94.2}    & \textbf{95.0} \\ \midrule
\multirow{5}{*}{UCF}          & 0.50                  & 70.5          & {\ul 71.4}    & 39.3 & 34.3 & 70.3       & 37.7 & 69.9 & 71.3          & \textbf{74.2} \\
                              & 0.75                  & {\ul 70.3}    & 68.8          & 42.3 & 31.2 & 65.2       & 33.0 & 58.3 & 69.0          & \textbf{73.1} \\
                              & 0.80                  & 65.3          & 63.3          & 37.9 & 37.2 & 65.6       & 34.9 & 59.3 & {\ul 68.5}    & \textbf{73.8} \\
                              & 0.88                  & 49.0          & 59.0          & 39.3 & 34.6 & 61.8       & 32.5 & 60.7 & {\ul 68.0}    & \textbf{72.3} \\
                              & 0.90                  & 59.3          & 59.6          & 34.7 & 24.3 & 59.2       & 32.5 & 54.3 & {\ul 66.8}    & \textbf{70.4} \\ \midrule
\multirow{5}{*}{FGVCAircraft} & 0.50                  & 28.4          & 29.6          & 17.2 & 14.7 & 28.4       & 12.5 & 27.4 & {\ul 34.9}    & \textbf{35.8} \\
                              & 0.75                  & 24.3          & 26.6          & 13.5 & 14.3 & 24.2       & 14.7 & 25.4 & {\ul 32.1}    & \textbf{32.9} \\
                              & 0.80                  & 23.7          & 25.1          & 13.7 & 11.5 & 23.2       & 11.6 & 24.5 & {\ul 32.2}    & \textbf{32.6} \\
                              & 0.88                  & 19.1          & 21.4          & 13.5 & 8.3  & 19.6       & 13.5 & 21.7 & \textbf{28.8} & {\ul 28.7}    \\
                              & 0.90                  & 17.6          & 22.2          & 13.8 & 10.5 & 17.8       & 11.5 & 21.4 & \textbf{28.1} & {\ul 26.6}    \\ \midrule
\multirow{5}{*}{OxfordPets}   & 0.50                  & \textbf{87.8} & 81.0          & 57.3 & 45.8 & 71.0       & 45.1 & 71.7 & 77.5          & {\ul 81.1}    \\
                              & 0.75                  & \textbf{84.8} & 80.3          & 54.0 & 41.7 & 62.0       & 37.9 & 64.4 & 73.9          & {\ul 80.5}    \\
                              & 0.80                  & \textbf{84.7} & 73.9          & 44.5 & 38.9 & 56.9       & 29.5 & 63.4 & 71.8          & {\ul 79.2}    \\
                              & 0.88                  & \textbf{80.8} & 67.2          & 41.0 & 34.6 & 44.4       & 25.5 & 47.6 & 67.2          & {\ul 71.9}    \\
                              & 0.90                  & \textbf{76.3} & 52.2          & 39.9 & 30.2 & 38.9       & 34.2 & 40.9 & 63.8          & {\ul 70.2}    \\ \bottomrule
\end{tabular}
\end{table}

\clearpage
In addition, we further conduct experiments on the fundus OCTA-3mm dataset to evaluate the effectiveness of our method in a medical image classification scenario under partial-label supervision. This dataset contains 200 OCTA images categorized into four ophthalmic classes: NORMAL, DR, AMD, and CNV. We construct partial-label annotations following a random candidate-label generation strategy, where each sample contains one true label and several randomly selected irrelevant labels. As shown in Table~\ref{tab:eye}, we report the classification accuracy of HopS when the number of candidate labels is set to 1, 2, 3, and 4, respectively.


\begin{table}[]
\centering
\caption{Testing accuracy (\%) of HopS on the OCTA-3mm dataset.}
\label{tab:eye}
\begin{tabular}{lcccc}
\toprule
Numbers of candidate labels & 1 & 2 & 3 & 4 \\
\midrule
Testing Accuracy & 81.80 & 84.80 & 42.40 & 9.10 \\
\bottomrule
\end{tabular}
\end{table}

\subsection{\textit{Insd}-Confusion Type.} Since partial-label losses perform well under the \textit{rand} confusion setting, we further compare HopS against them under instance-dependent partial labels. The \textit{Insd} candidate sets, constructed based on visual similarity, are derived from features extracted by pre-trained backbones, including ResNet-18 (R18), ResNet-50 (R50), and CLIP-ResNet50 (CLIP). As shown in Tables~\ref{tab:insd_1} and ~\ref{tab:insd_2}, HopS consistently demonstrates significant advantages.

\begin{table}[ht]
\centering
\caption{Testing accuracy (\%) under \textit{uni}-prompt with \textit{insd} confusion rates $\gamma_c$=0.80.}
\label{tab:insd_1}
\resizebox{\columnwidth}{!}{
\begin{tabular}{@{\hskip 3pt}c||c||c|c|c|c|c|c||c||c|c|c|c|c|c@{\hskip 3pt}}
\toprule
Dataset                  & Insd & CC            & CE            & EXP  & GCE  & LWC           & HopS                & Dataset                       & CC         & CE                  & EXP  & GCE  & LWC        & HopS                \\ \midrule
\multirow{3}{*}{Caltech} & CLIP & \textbf{46.2} & 41.7          & 4.2  & 19.6 & 39.4          & {\ul 45.0}          & \multirow{3}{*}{Flowers}      & {\ul 40.6} & 31.9                & 15.4 & 13.6 & {\ul 38.6} & \textbf{59.2}       \\
                         & R18  & 25.4          & 28.1          & 6.1  & 0.5  & {\ul 33.9}    & \textbf{49.5}       &                               & {\ul 36.0} & 23.7                & 15.8 & 8.8  & {\ul 28.5} & \textbf{69.9}       \\
                         & R50  & 23.4          & {\ul 37.0}    & 25.4 & 10.0 & 31.6          & \textbf{38.9}       &                               & {\ul 45.4} & 26.5                & 11.5 & 12.1 & {\ul 32.1} & \textbf{65.0}       \\ \midrule
\multirow{3}{*}{DTD}     & CLIP & 25.7          & 30.9          & 19.4 & 7.6  & \textbf{38.1} & {\ul 37.4}          & \multirow{3}{*}{OxfordPets}    & {\ul 36.1} & 42.3                & 22.0 & 17.3 & {\ul 44.7} & \textbf{48.8}       \\
                         & R18  & 32.2          & 31.2          & 19.1 & 16.6 & {\ul 37.8}    & \textbf{43.1}       &                               & {\ul 35.3} & {\ul 47.5}          & 27.0 & 16.0 & 47.0       & \textbf{55.6}       \\
                         & R50  & 25.8          & 32.0          & 16.7 & 15.1 & {\ul 34.5}    & \textbf{42.5}       &                               & {\ul 29.9} & 44.6                & 28.8 & 18.3 & {\ul 45.0} & \textbf{51.5}       \\ \midrule
\multirow{3}{*}{Food}    & CLIP & 33.7          & {\ul 27.2}    & 14.5 & 15.8 & {\ul 37.8}    & \textbf{49.0}       & \multirow{3}{*}{EuroSAT}      & 15.8       & {\ul 38.9}          & 11.8 & 11.4 & {\ul 32.5} & \textbf{43.5}       \\
                         & R18  & 39.0          & \textbf{41.2} & 15.4 & 13.2 & {\ul 42.1}    & {\ul \textbf{56.5}} &                               & 14.8       & {\ul \textbf{41.0}} & 12.0 & 18.9 & 28.9       & {\ul \textbf{40.0}} \\
                         & R50  & 42.3          & {\ul 43.6}    & 13.2 & 13.7 & {\ul 44.0}    & \textbf{53.7}       &                               & 16.0       & {\ul 44.4}          & 18.5 & 9.3  & {\ul 27.0} & \textbf{44.9}       \\ \midrule
\multirow{3}{*}{UCF}     & CLIP & {\ul 27.7}    & 21.9          & 14.6 & 10.9 & {\ul 31.3}    & \textbf{33.7}       & \multirow{3}{*}{FGVCAircraft} & {\ul 11.7} & 11.4                & 8.2  & 6.7  & {\ul 11.5} & \textbf{13.0}       \\
                         & R18  & {\ul 31.8}    & 32.4          & 11.6 & 11.5 & {\ul 37.7}    & \textbf{53.2}       &                               & {\ul 11.7} & 8.2                 & 6.3  & 4.4  & {\ul 8.2}  & \textbf{13.7}       \\
                         & R50  & {\ul 27.8}    & {\ul 31.3}    & 12.5 & 7.6  & 29.2          & \textbf{43.0}       &                               & {\ul 11.5} & {\ul 10.0}          & 5.2  & 2.3  & 9.5        & \textbf{13.3} \\ \bottomrule
\end{tabular}
}
\end{table}

\begin{table}[ht]
\centering
\caption{Testing accuracy (\%) under \textit{uni}-prompt across two \textit{insd} confusion rates.}
\label{tab:insd_2}
\resizebox{\columnwidth}{!}{
\begin{tabular}{@{\hskip 3pt}c||c||c|c|c|c|c|c||c|c|c|c|c|c@{\hskip 3pt}}
\toprule
    \multirow[c]{2}{*}{Dataset} & \multirow[c]{2}{*}{\textit{Insd}}
      & \multicolumn{6}{c||}{$\gamma_c$=0.67} & \multicolumn{6}{c}{$\gamma_c$=0.75} \\
    \cmidrule(lr){3-14}
     & & CC & CE & EXP & GCE & LWC & HopS & CC & CE & EXP & GCE & LWC & HopS \\
    \midrule
\multirow{3}{*}{Caltech}      & CLIP             & \textbf{57.7} & 49.2       & 24.9 & 13.3 & 49.5       & {\ul 57.6}    & 41.2          & 46.8 & 20.4 & 9.5  & \textbf{52.3} & {\ul 49.0}    \\ 
                              & R18                  & 41.0          & {\ul 54.7} & 29.2 & 12.8 & 54.3       & \textbf{64.1} & {\ul 46.2}    & 45.5 & 17.1 & 5.6  & 44.7          & \textbf{58.2} \\
                              & R50                  & 28.4          & 42.2       & 12.3 & 8.6  & {\ul 49.5} & \textbf{50.3} & 31.2          & 35.9 & 18.7 & 9.2  & {\ul 38.7}    & \textbf{45.3} \\ \midrule
\multirow{3}{*}{DTD}          & CLIP             & 39.4          & 40.4       & 24.7 & 25.2 & {\ul 44.7} & \textbf{47.8} & 28.7          & 35.5 & 20.6 & 12.8 & \textbf{41.6} & {\ul 41.4}    \\
                              & R18                  & {\ul 47.3}    & 42.8       & 19.7 & 20.0 & 46.4       & \textbf{48.8} & 33.7          & 40.7 & 12.7 & 20.8 & {\ul 40.7}    & \textbf{47.3} \\
                              & R50                  & 37.1          & 42.1       & 19.1 & 17.7 & {\ul 44.3} & \textbf{51.7} & 31.5          & 37.1 & 17.3 & 23.6 & {\ul 40.3}    & \textbf{46.2} \\ \midrule
\multirow{3}{*}{EuroSAT}      & CLIP             & 32.8          & {\ul 37.6} & 33.6 & 24.0 & 36.7       & \textbf{57.5} & 15.5          & 21.0 & 15.3 & 11.5 & {\ul 27.5}    & \textbf{45.8} \\
                              & R18                  & 34.4          & 35.3       & 29.3 & 35.0 & {\ul 37.3} & \textbf{48.5} & 23.4          & 30.1 & 15.1 & 17.3 & {\ul 33.5}    & \textbf{49.8} \\
                              & R50                  & 37.2          & 35.8       & 32.0 & 20.0 & {\ul 39.6} & \textbf{52.7} & 22.0          & 31.7 & 27.4 & 16.4 & {\ul 37.0}    & \textbf{37.3} \\ \midrule
\multirow{3}{*}{FGVCAircraft} & CLIP             & {\ul 15.7}    & 12.2       & 8.3  & 6.2  & 12.5       & \textbf{17.9} & \textbf{14.9} & 11.2 & 9.3  & 7.1  & 12.7          & {\ul 14.0}    \\
                              & R18                  & \textbf{14.6} & 10.0       & 7.8  & 5.0  & 9.5        & {\ul 13.9}    & \textbf{13.1} & 8.3  & 5.4  & 6.3  & 8.9           & {\ul 13.1}    \\
                              & R50                  & {\ul 13.7}    & 10.4       & 7.5  & 5.1  & 10.3       & \textbf{14.3} & \textbf{12.8} & 8.9  & 4.9  & 5.0  & 9.4           & {\ul 12.6}    \\ \midrule
\multirow{3}{*}{Food}         & CLIP             & 40.2          & 45.0       & 23.2 & 18.8 & {\ul 47.1} & \textbf{54.7} & 35.8          & 39.2 & 15.3 & 13.5 & {\ul 46.4}    & \textbf{48.9} \\
                              & R18                  & {\ul 54.4}    & 53.4       & 17.0 & 18.2 & 48.5       & \textbf{59.7} & {\ul 47.9}    & 44.1 & 13.6 & 12.3 & 45.0          & \textbf{59.4} \\
                              & R50                  & {\ul 53.8}    & 47.1       & 16.1 & 12.3 & 50.6       & \textbf{62.5} & 43.6          & 35.6 & 16.7 & 13.3 & {\ul 46.0}    & \textbf{57.0} \\ \midrule
\multirow{3}{*}{Flowers}      & CLIP             & {\ul 64.6}    & 32.6       & 18.4 & 20.4 & 48.4       & \textbf{73.9} & {\ul 41.8}    & 40.5 & 22.0 & 11.4 & 37.2          & \textbf{64.8} \\
                              & R18                  & {\ul 68.5}    & 29.6       & 21.4 & 19.6 & 42.5       & \textbf{82.5} & {\ul 47.5}    & 29.7 & 20.8 & 14.5 & 35.9          & \textbf{75.1} \\
                              & R50                  & {\ul 72.9}    & 44.9       & 22.3 & 11.6 & 37.7       & \textbf{82.0} & {\ul 53.3}    & 21.0 & 17.8 & 20.0 & 36.1          & \textbf{74.0} \\ \midrule
\multirow{3}{*}{OxfordPets}    & CLIP             & {\ul 47.4}    & 40.8       & 23.0 & 23.2 & 47.3       & \textbf{58.5} & 33.3          & 37.8 & 16.5 & 22.6 & {\ul 41.3}    & \textbf{50.3} \\
                              & R18                  & {\ul 49.6}    & 41.3       & 32.8 & 18.7 & 48.5       & \textbf{59.1} & 42.5          & 44.3 & 26.3 & 12.7 & {\ul 46.1}    & \textbf{55.6} \\
                              & R50                  & {\ul 44.1}    & 36.7       & 25.3 & 28.0 & 41.0       & \textbf{53.3} & 33.8          & 35.9 & 18.5 & 19.2 & {\ul 38.2}    & \textbf{51.8} \\ \midrule
\multirow{3}{*}{UCF}          & CLIP             & {\ul 41.2}    & 36.8       & 17.6 & 14.3 & 36.4       & \textbf{49.3} & {\ul 35.0}    & 29.1 & 14.4 & 12.9 & 34.6          & \textbf{38.4} \\
                              & R18                  & {\ul 52.4}    & 41.2       & 15.4 & 15.0 & 49.0       & \textbf{60.9} & {\ul 43.2}    & 32.9 & 22.2 & 13.1 & 40.3          & \textbf{53.5} \\
                              & R50                  & {\ul 48.7}    & 39.4       & 9.1  & 14.5 & 44.1       & \textbf{59.7} & {\ul 40.8}    & 30.1 & 12.8 & 9.0  & 36.0          & \textbf{49.1} \\ \bottomrule
\end{tabular}}
\end{table}

\section{Further Analysis}\label{fa}
\subsection{Mutual Complementation.} 
\begin{figure}[h]
    \centering
    \includegraphics[width=0.75\linewidth]{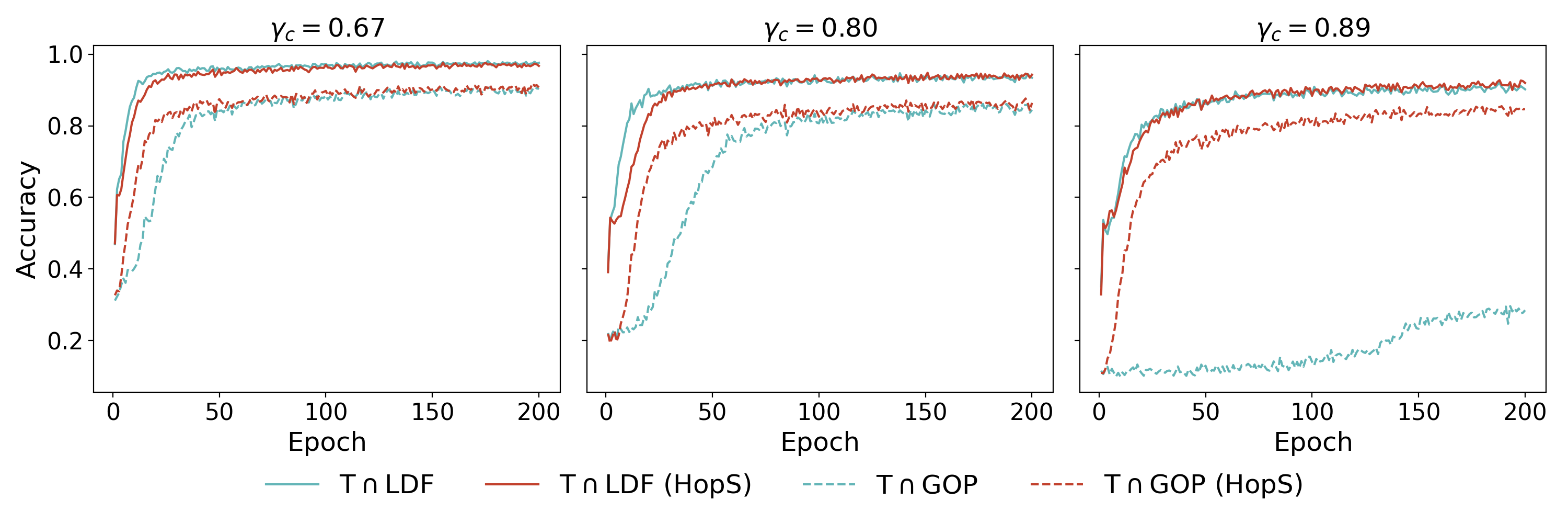}
    \caption{Identifying accuracy of LDF, GOP, and HopS on the Caltech.}
    \label{fig:cal_3}
\end{figure}

\begin{figure}[h]
    \centering
    \includegraphics[width=0.75\linewidth]{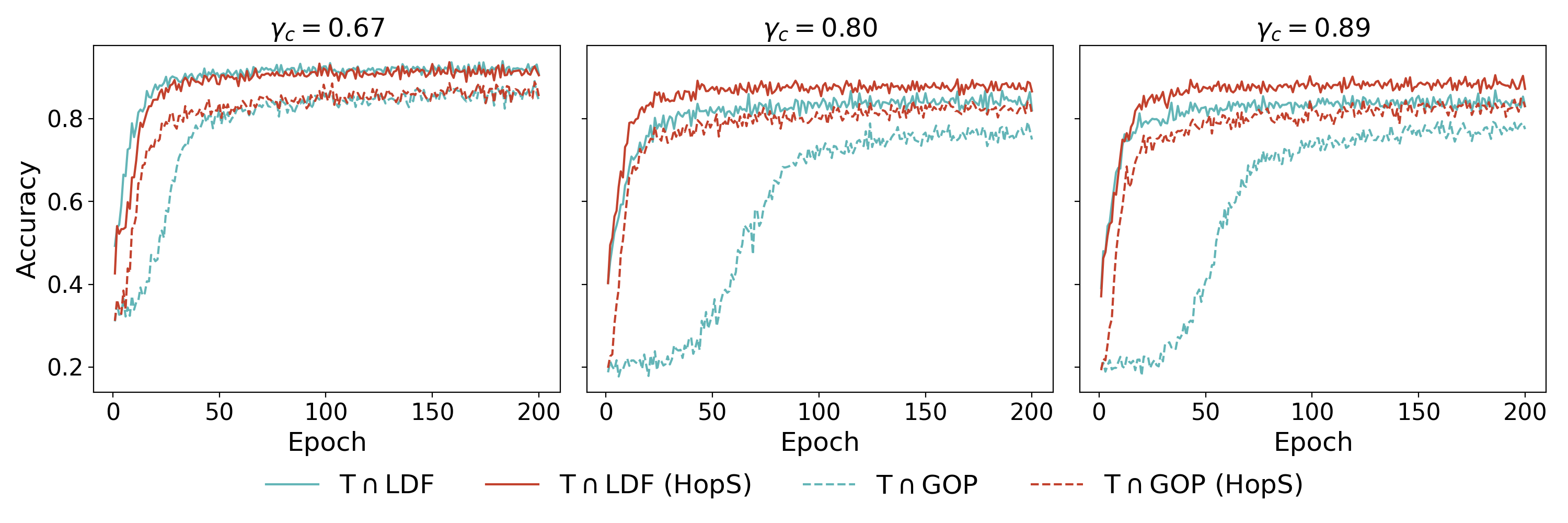}
    \caption{Identifying accuracy of LDF, GOP, and HopSon the DTD.}
    \label{fig:dtd_3}
\end{figure}

\begin{figure}[h]
    \centering
    \includegraphics[width=0.75\linewidth]{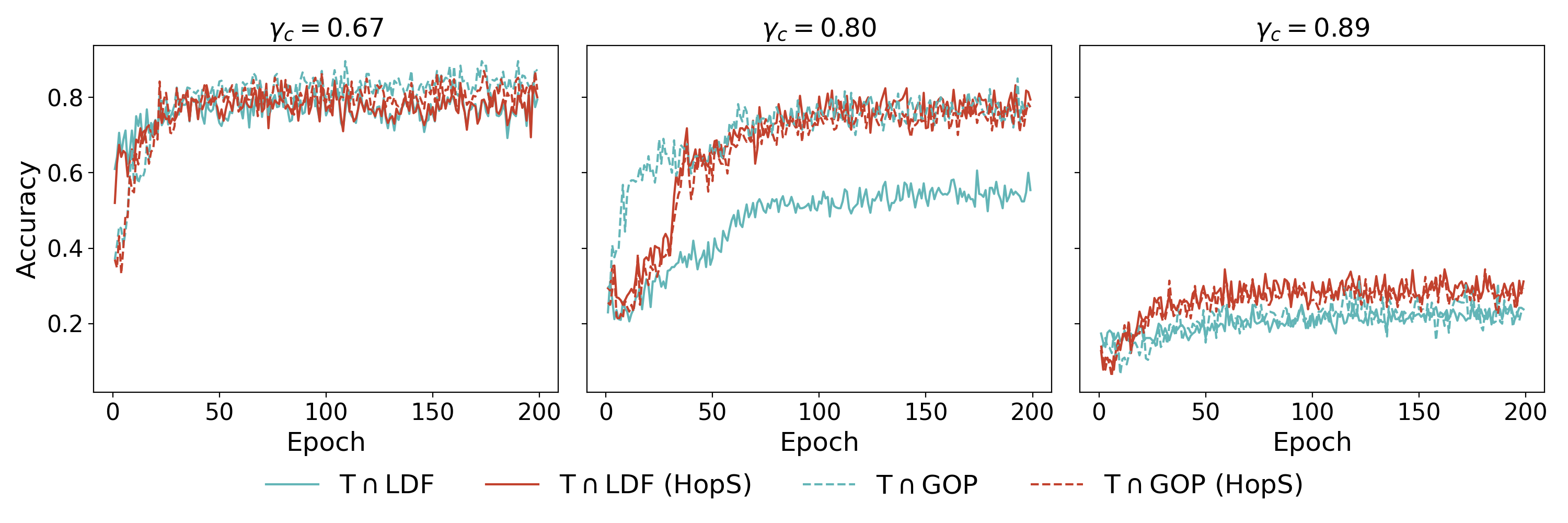}
    \caption{Identifying accuracy of LDF, GOP, and HopS on the EuroSAT.}
    \label{fig:eur_3}
\end{figure}

\begin{figure}[H]
    \centering
    \includegraphics[width=0.75\linewidth]{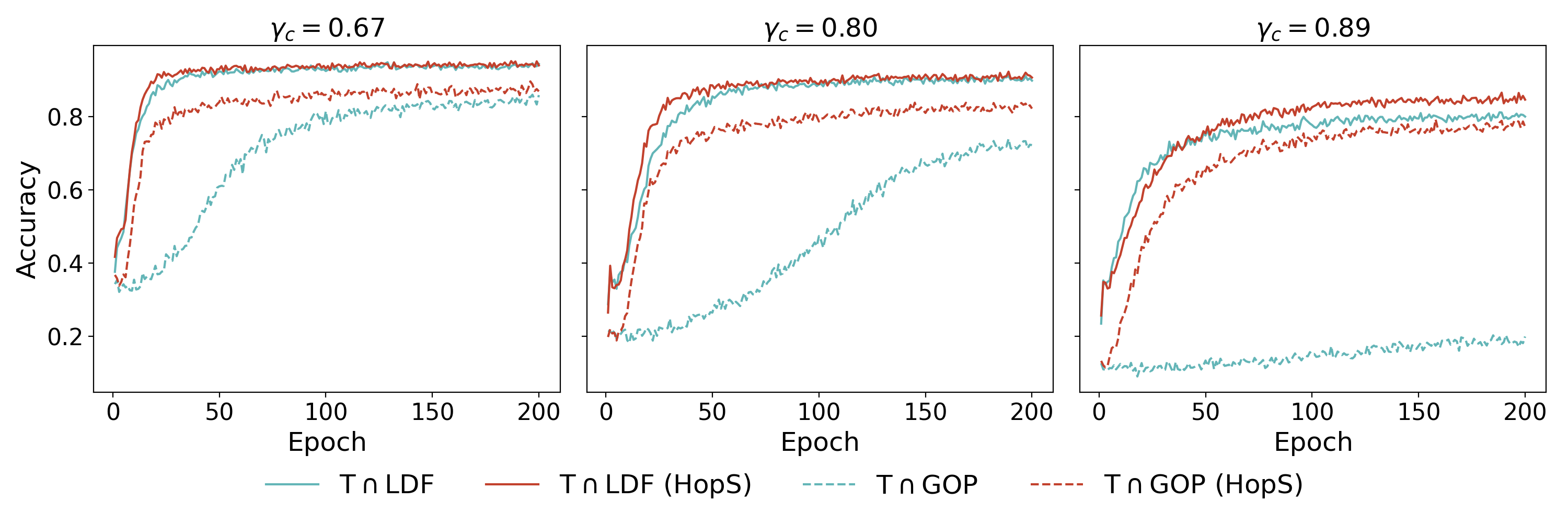}
    \caption{Identifying accuracy of LDF, GOP, and HopS on the Food.}
    \label{fig:food_3}
\end{figure}

\begin{figure}[H]
    \centering
    \includegraphics[width=0.75\linewidth]{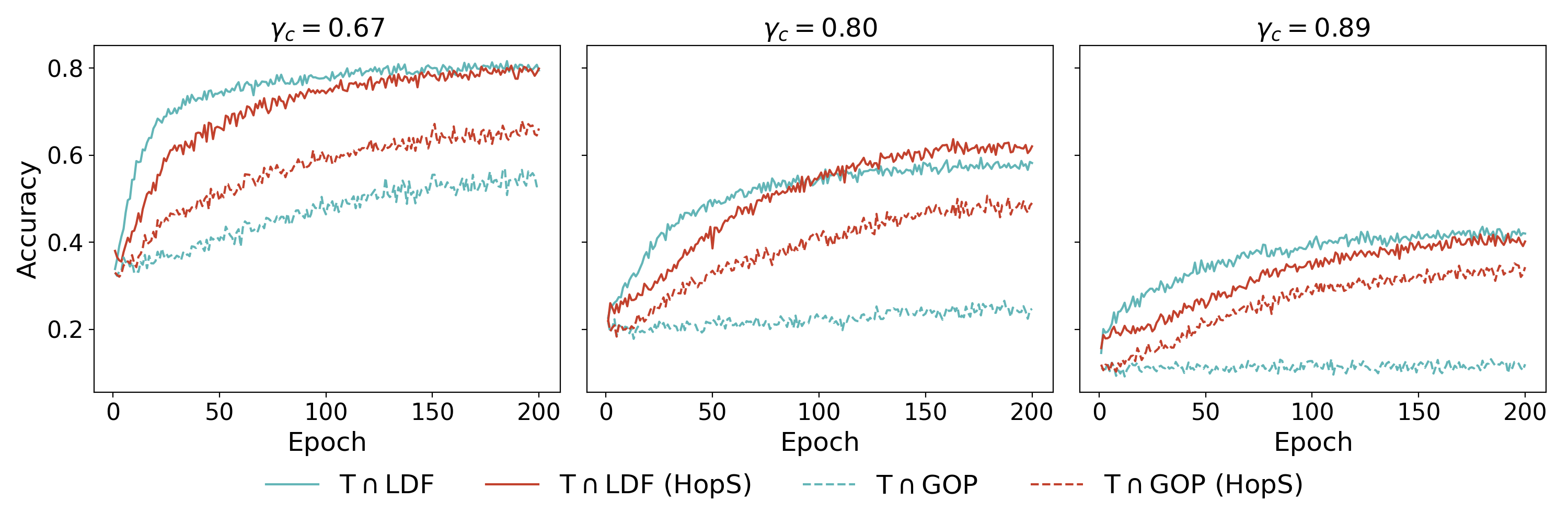}
    \caption{Identifying accuracy of LDF, GOP, and HopS on the FGVCAircraft.}
    \label{fig:fa_3}
\end{figure}

\begin{figure}[H]
    \centering
    \includegraphics[width=0.75\linewidth]{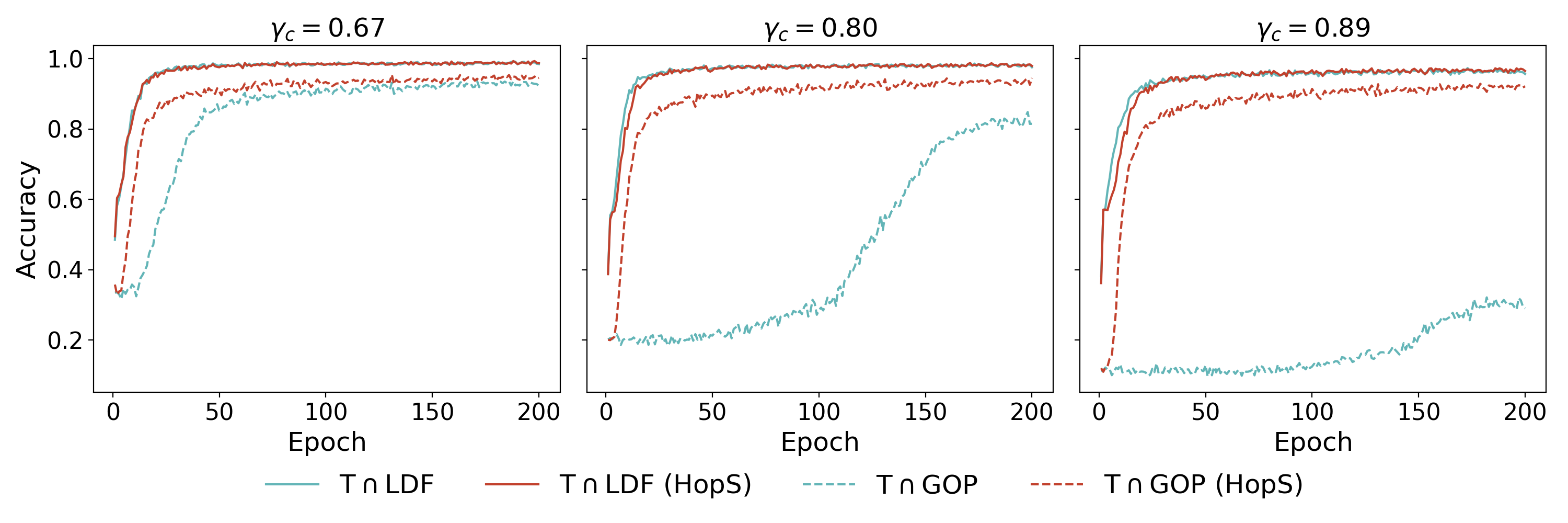}
    \caption{Identifying accuracy of LDF, GOP, and HopS on the Flowers.}
    \label{fig:fl_3}
\end{figure}

\begin{figure}[H]
    \centering
    \includegraphics[width=0.75\linewidth]{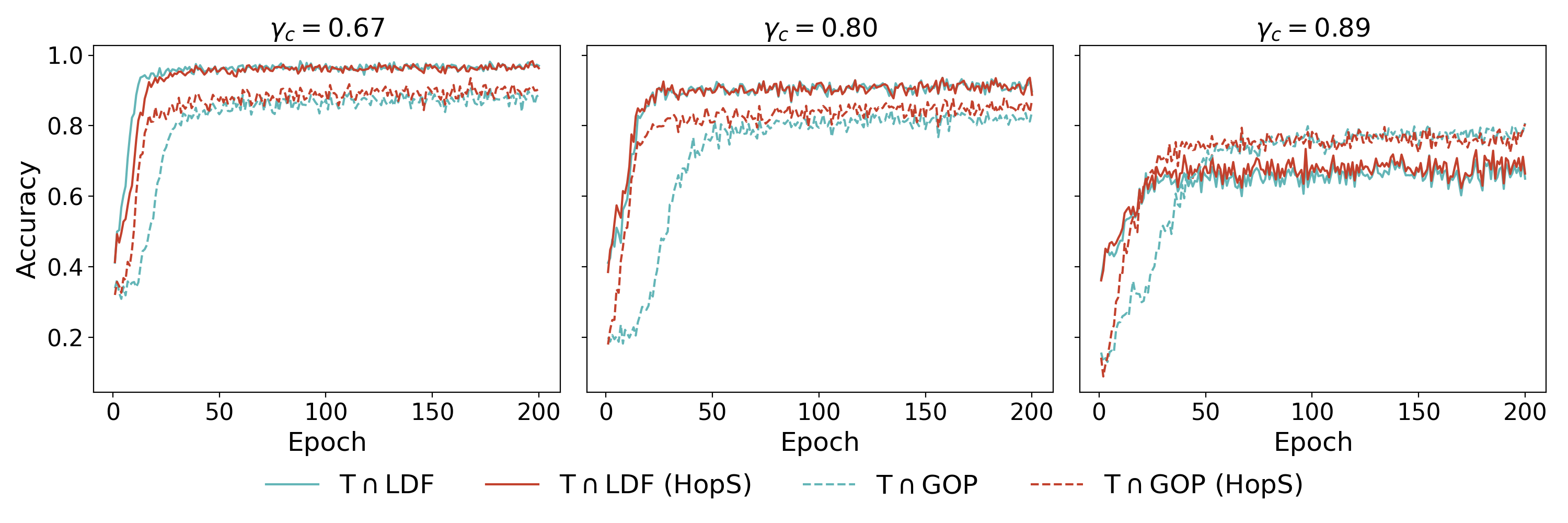}
    \caption{Identifying accuracy of LDF, GOP, and HopS on the OxfordPets.}
    \label{fig:ox_3}
\end{figure}

\begin{figure}[H]
    \centering
    \includegraphics[width=0.75\linewidth]{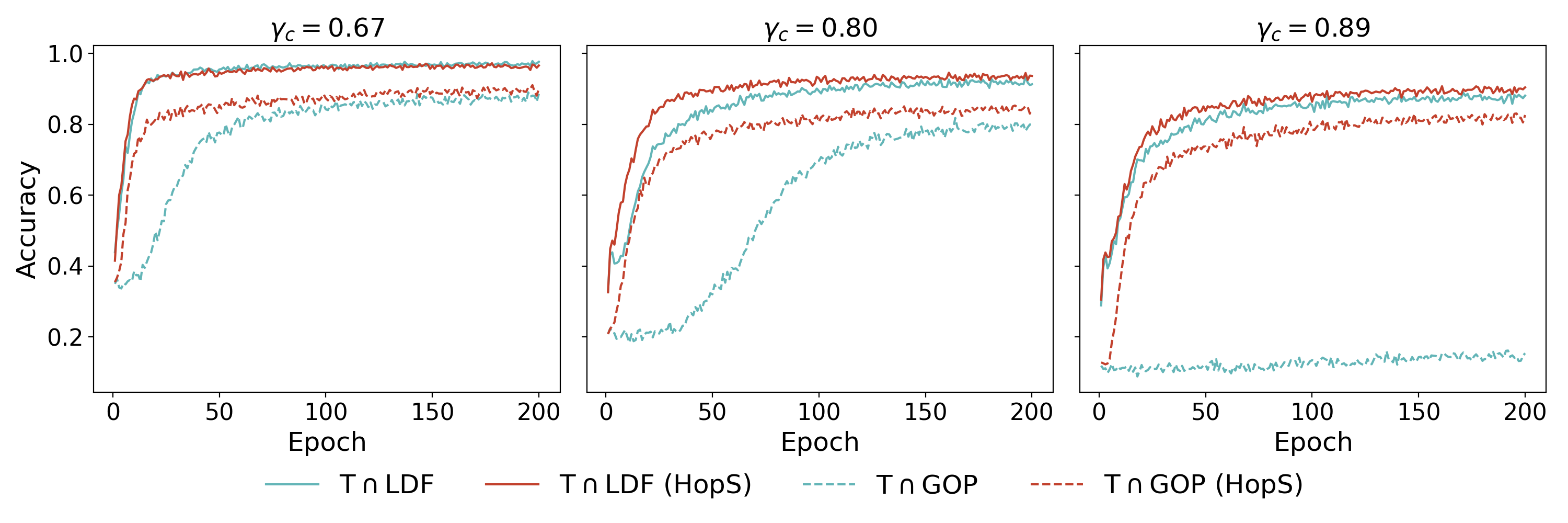}
    \caption{Identifying accuracy of LDF, GOP, and HopS on the UCF.}
    \label{fig:ucf_3}
\end{figure}

To evaluate the interaction between the LDF and GOP modules, we compare their individual performance with their performance within the overall HopS model in terms of identifying the ground-truth labels \textbf{T}, considering both the training and testing phases. \textbf{Training}. Figures~\ref{fig:cal_3} -- \ref{fig:ucf_3} illustrates the identifying accuracy of LDF, GOP, and HopS over the epochs, highlighting the role of LDF in accelerating early-stage optimization by guiding the global module. Moreover, the proportion of correctly identified labels by each module within HopS ultimately surpasses their standalone performance. \textbf{Testing}. Table~\ref{tab:abs} demonstrates that a weighting coefficient of $\lambda = 1.0$ achieves a proper balance between local and global signals, thereby consistently improving performance. Conversely, under high confusion, overemphasizing either component results in performance degradation, highlighting the necessity of their complementary interaction.


\begin{table}[H]
\centering
\caption{Testing phases: (a) module validity, (b) \(\lambda\) sensitivity.}
\label{tab:abs}
\begin{adjustbox}{angle=90,center}

\begin{minipage}{0.9\textheight}
\centering

\begin{minipage}{0.48\textwidth}
\centering
\text{(a) Testing accuracy (\%) under the \textit{uni}(left) and \textit{cls}(right).} \\[0.5em]
\begin{tabular}{@{\hskip 5pt}c||c||c|c|c||c|c|c@{\hskip 5pt}}
\toprule
Dataset & \(\gamma_c\) & HopS & GOP & LDF & HopS & GOP & LDF \\ \midrule
\multirow{3}{*}{Caltech} 
& 0.67 & 88.6 & \textbf{89.7} & 88.8 & \textbf{90.7} & 90.5 & 88.9 \\
& 0.80 & \textbf{89.0} & 88.8 & 88.0 & 90.3 & \textbf{91.0} & 88.8 \\
& 0.89 & \textbf{88.7} & 56.9 & 86.9 & 90.2 & \textbf{90.8} & 88.5 \\ \midrule

\multirow{3}{*}{DTD} 
& 0.67 & 59.1 & \textbf{60.0} & 59.5 & \textbf{61.7} & 59.8 & 58.0 \\
& 0.80 & \textbf{59.0} & 56.3 & 55.2 & \textbf{58.9} & 58.5 & 55.2 \\
& 0.89 & \textbf{60.0} & 56.3 & 54.5 & 58.3 & \textbf{58.8} & 56.8 \\ \midrule

\multirow{3}{*}{EuroSAT} 
& 0.67 & 77.2 & \textbf{78.5} & 74.0 & 78.9 & \textbf{80.8} & 76.5 \\
& 0.80 & \textbf{75.3} & 74.9 & 54.1 & \textbf{79.0} & 75.7 & 66.8 \\
& 0.89 & \textbf{44.1} & 33.5 & 18.6 & \textbf{52.9} & 42.5 & 28.3 \\ \midrule

\multirow{3}{*}{FGVCAircraft} 
& 0.67 & 23.7 & 19.9 & \textbf{27.0} & \textbf{34.9} & 33.8 & 29.8 \\
& 0.80 & \textbf{22.2} & 11.1 & 21.4 & \textbf{32.1} & 31.1 & 25.9 \\
& 0.89 & \textbf{18.3} & 3.0 & 17.5 & \textbf{28.3} & 24.3 & 23.6 \\ \midrule

\multirow{3}{*}{Food} 
& 0.67 & \textbf{67.5} & 61.6 & 66.3 & 68.8 & \textbf{68.9} & 65.8 \\
& 0.80 & 66.4 & 53.3 & \textbf{67.0} & 68.4 & \textbf{68.7} & 62.9 \\
& 0.89 & \textbf{63.9} & 13.4 & 62.4 & 66.6 & \textbf{67.5} & 63.3 \\ \midrule

\multirow{3}{*}{Flowers} 
& 0.67 & 92.5 & 91.2 & \textbf{92.6} & 95.3 & \textbf{95.5} & 94.8 \\
& 0.80 & \textbf{92.4} & 85.4 & 92.3 & \textbf{95.1} & \textbf{95.1} & 93.9 \\
& 0.89 & \textbf{93.2} & 29.5 & 91.5 & \textbf{95.3} & 95.0 & 93.8 \\ \midrule

\multirow{3}{*}{OxfordPets} 
& 0.67 & 83.2 & \textbf{83.5} & 81.8 & 79.8 & \textbf{80.4} & 79.7 \\
& 0.80 & \textbf{81.6} & 81.0 & \textbf{81.6} & 78.1 & \textbf{79.1} & 76.2 \\
& 0.89 & 76.6 & \textbf{81.7} & 71.1 & 68.4 & \textbf{77.7} & 62.7 \\ \midrule

\multirow{3}{*}{UCF} 
& 0.67 & 70.3 & 69.8 & \textbf{71.7} & \textbf{72.8} & 72.4 & 70.2 \\
& 0.80 & \textbf{69.9} & 67.4 & 68.0 & \textbf{72.3} & 71.5 & 69.4 \\
& 0.89 & \textbf{69.0} & 11.8 & 67.5 & 71.0 & \textbf{71.1} & 70.0 \\ \bottomrule
\end{tabular}
\end{minipage}
\hfill
\begin{minipage}{0.48\textwidth}
\centering
\text{(b) Testing accuracy (\%) across a wide range of $\lambda$.} \\[0.5em]
\begin{tabular}{@{\hskip 5pt}c||c||c|c|c|c|c@{\hskip 5pt}}
\toprule
Dataset & \(\lambda\) & 0.50 & 0.75 & 0.80 & 0.88 & 0.90 \\ \midrule
\multirow{3}{*}{Caltech} & 1.0 & \textbf{89.6} & 90.1 & 89.5 & \textbf{90.3} & \textbf{89.8} \\
& 2.0 & 89.2 & \textbf{90.7} & \textbf{89.8} & 88.8 & 89.4 \\
& 0.5 & 89.4 & 89.5 & 89.2 & 88.8 & 88.2 \\ \midrule
\multirow{3}{*}{DTD} & 1.0 & \textbf{63.2} & 59.3 & \textbf{60.2} & 58.2 & \textbf{56.3} \\
& 2.0 & 62.9 & \textbf{60.3} & 58.9 & \textbf{58.4} & 54.1 \\
& 0.5 & 62.2 & 58.3 & 57.9 & 56.9 & 53.7 \\ \midrule
\multirow{3}{*}{EuroSAT} & 1.0 & \textbf{82.7} & 80.2 & \textbf{74.5} & \textbf{69.6} & \textbf{34.6} \\
& 2.0 & 82.5 & \textbf{81.9} & 73.3 & 55.5 & 34.4 \\
& 0.5 & 82.2 & 80.8 & 70.8 & 46.3 & 31.5 \\ \midrule
\multirow{3}{*}{FGVCAircraft} & 1.0 & \textbf{27.4} & 22.9 & 21.4 & \textbf{20.2} & 15.2 \\
& 2.0 & 25.4 & 23.3 & 21.5 & 19.2 & 14.0 \\
& 0.5 & 27.4 & \textbf{24.2} & \textbf{26.1} & 19.7 & \textbf{17.9} \\ \midrule
\multirow{3}{*}{Food} & 1.0 & 69.6 & 68.3 & 66.6 & 65.2 & 63.7 \\
& 2.0 & \textbf{69.9} & 65.5 & 63.9 & \textbf{66.8} & \textbf{67.2} \\
& 0.5 & 68.4 & \textbf{69.2} & \textbf{68.9} & 64.1 & 64.1 \\ \midrule
\multirow{3}{*}{Flowers} & 1.0 & \textbf{93.2} & \textbf{93.5} & 91.8 & \textbf{92.9} & 92.4 \\
& 2.0 & 90.1 & 92.3 & \textbf{93.5} & 89.2 & \textbf{93.5} \\
& 0.5 & 90.8 & 92.5 & 87.9 & 91.8 & 92.6 \\ \midrule
\multirow{3}{*}{OxfordPets} & 1.0 & 83.4 & 82.7 & 83.7 & 79.7 & 80.0 \\
& 2.0 & \textbf{83.5} & 79.9 & \textbf{84.6} & \textbf{81.6} & \textbf{81.3} \\
& 0.5 & 82.8 & \textbf{83.3} & 84.0 & 76.0 & 76.2 \\ \midrule
\multirow{3}{*}{UCF} & 1.0 & 71.1 & 71.5 & \textbf{71.5} & \textbf{69.3} & 67.4 \\
& 2.0 & \textbf{71.4} & \textbf{71.9} & 67.8 & 67.9 & \textbf{69.0} \\
& 0.5 & 71.0 & 71.9 & 69.6 & 67.3 & 69.0 \\ \bottomrule
\end{tabular}
\end{minipage}
\end{minipage}
\end{adjustbox}
\end{table}

\subsection{Effect of Batch Size.} 
\begin{figure}[H]
    \centering
    \includegraphics[width=\linewidth]{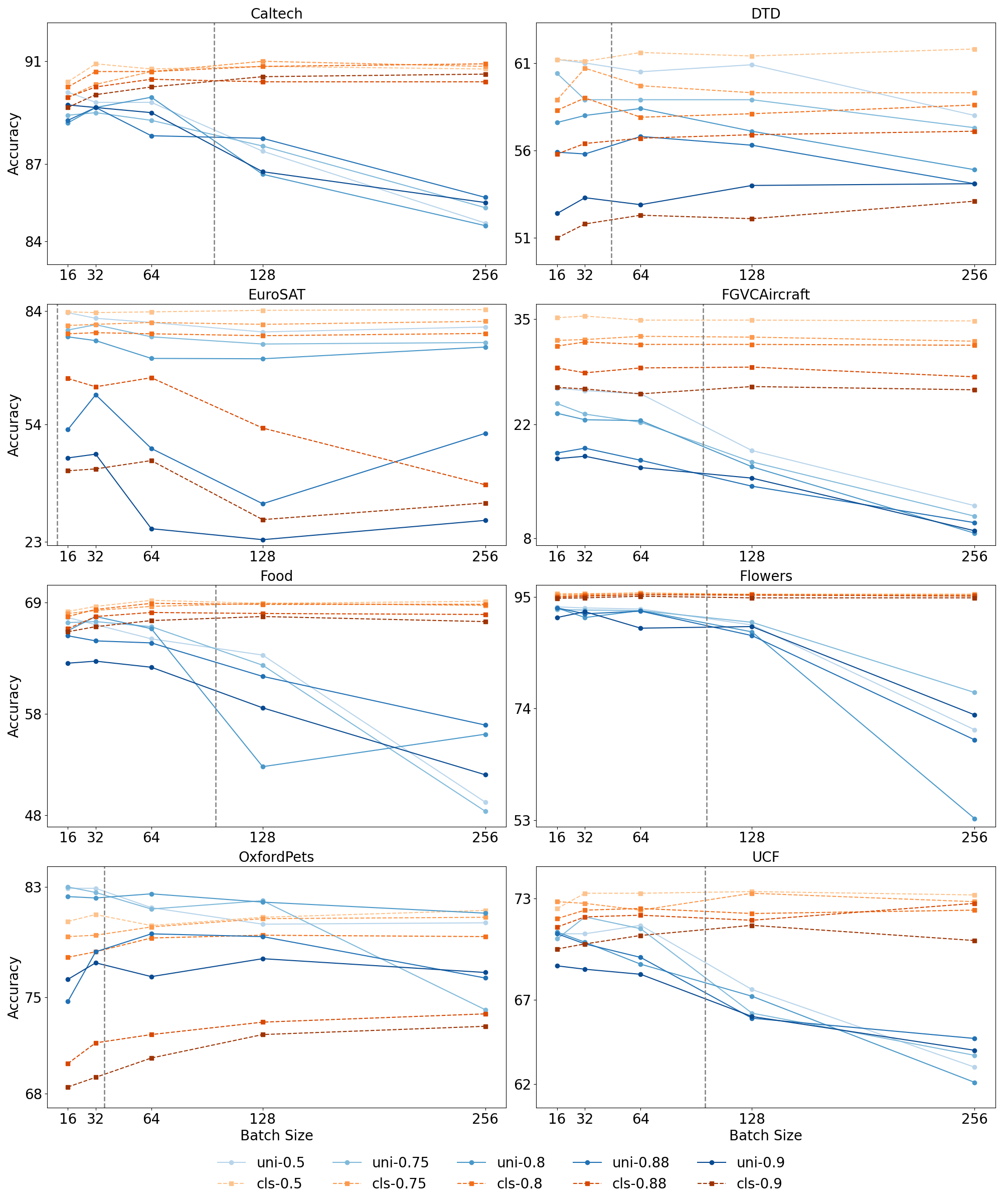}
    \caption{Testing accuracy of the HopS across different \(B\) under five confusion rates.}
    \label{fig:bs}
\end{figure}

We investigate the effect of varying batch sizes \(B \in \{16, 32, 64, 128, 256\}\) on the performance of HopS, focusing on the testing accuracy under varying levels of label confusion \(\gamma_c \in\ \{0.5, 0.75, 0.80, 0.88, 0.90\}\). As shown in Figure~\ref{fig:bs} and Table~\ref{tab:bs}, the results compare the performance under \textit{uni}-prompt and \textit{cls}-prompts.

\begin{table}[h]
\centering
\caption{Testing accuracy (\%) of HopS under \textit{uni} (left) and \textit{cls} (right) prompts.}
\label{tab:bs}
\resizebox{0.85\linewidth}{!}{
\begin{tabular}{@{}c||c||c|c|c|c|c||c|c|c|c|c@{}}
\toprule
Dataset                       & \(\gamma_c\) & 16  & 32  & 64  & 128 & 256 & 16  & 32  & 64  & 128 & 256 \\ \midrule
\multirow{5}{*}{Caltech}   & 0.50                  & 89.8 & 89.4 & 89.4 & 87.5 & 84.7 & 90.2 & 90.9 & 90.7 & 90.8 & 90.7 \\
                              & 0.75                  & 88.9 & 89.0 & 88.7 & 87.7 & 85.3 & 89.6 & 90.1 & 90.6 & 91.0 & 90.8 \\
                              & 0.80                  & 88.6 & 89.2 & 89.6 & 86.6 & 84.6 & 90.0 & 90.6 & 90.6 & 90.8 & 90.9 \\
                              & 0.88                  & 88.7 & 89.2 & 88.1 & 88.0 & 85.7 & 89.6 & 90.0 & 90.3 & 90.2 & 90.2 \\
                              & 0.90                  & 89.3 & 89.2 & 89.0 & 86.7 & 85.5 & 89.2 & 89.7 & 90.0 & 90.4 & 90.5 \\ \midrule
\multirow{5}{*}{EuroSAT}      & 0.50                  & 61.2 & 61.0 & 60.5 & 60.9 & 58.0 & 61.2 & 61.1 & 61.6 & 61.4 & 61.8 \\
                              & 0.75                  & 60.4 & 58.9 & 58.9 & 58.9 & 57.3 & 58.9 & 60.7 & 59.7 & 59.3 & 59.3 \\
                              & 0.80                  & 57.6 & 58.0 & 58.4 & 57.1 & 54.9 & 58.3 & 59.0 & 57.9 & 58.1 & 58.6 \\
                              & 0.88                  & 55.9 & 55.8 & 56.8 & 56.3 & 54.1 & 55.8 & 56.4 & 56.7 & 56.9 & 57.1 \\
                              & 0.90                  & 52.4 & 53.3 & 52.9 & 54.0 & 54.1 & 51.0 & 51.8 & 52.3 & 52.1 & 53.1 \\ \midrule
\multirow{5}{*}{UCF}       & 0.50                  & 83.6 & 82.1 & 81.0 & 78.5 & 79.8 & 83.8 & 83.6 & 83.8 & 84.2 & 84.4 \\
                              & 0.75                  & 79.0 & 80.4 & 77.2 & 75.3 & 75.7 & 80.2 & 80.5 & 81.0 & 80.5 & 81.3 \\
                              & 0.80                  & 77.2 & 76.2 & 71.5 & 71.4 & 74.5 & 78.0 & 78.3 & 78.0 & 77.5 & 78.1 \\
                              & 0.88                  & 52.7 & 61.9 & 47.7 & 33.1 & 51.7 & 66.2 & 64.0 & 66.4 & 53.1 & 38.1 \\
                              & 0.90                  & 45.2 & 46.2 & 26.5 & 23.6 & 28.7 & 41.8 & 42.3 & 44.5 & 28.9 & 33.3 \\ \midrule
\multirow{5}{*}{Flowers}   & 0.50                  & 26.5 & 26.2 & 25.8 & 18.8 & 12.0 & 35.2 & 35.4 & 34.9 & 34.9 & 34.8 \\
                              & 0.75                  & 24.6 & 23.3 & 22.3 & 17.4 & 10.7 & 32.4 & 32.5 & 32.9 & 32.8 & 32.3 \\
                              & 0.80                  & 23.4 & 22.6 & 22.5 & 16.8 & 8.6  & 31.7 & 32.2 & 31.9 & 31.9 & 31.8 \\
                              & 0.88                  & 18.5 & 19.1 & 17.6 & 14.4 & 9.9  & 29.0 & 28.4 & 29.0 & 29.1 & 27.9 \\
                              & 0.90                  & 17.8 & 18.1 & 16.7 & 15.4 & 8.9  & 26.6 & 26.4 & 25.8 & 26.7 & 26.3 \\ \midrule
\multirow{5}{*}{Food}      & 0.50                  & 67.5 & 66.8 & 65.4 & 63.8 & 49.3 & 68.1 & 68.6 & 69.2 & 68.9 & 69.1 \\
                              & 0.75                  & 67.0 & 67.1 & 66.6 & 62.8 & 48.4 & 67.9 & 68.2 & 68.6 & 68.9 & 68.7 \\
                              & 0.80                  & 66.2 & 67.6 & 66.4 & 52.8 & 56.0 & 67.6 & 68.3 & 68.9 & 68.8 & 68.8 \\
                              & 0.88                  & 65.7 & 65.2 & 65.0 & 61.7 & 56.9 & 66.4 & 67.6 & 68.0 & 67.9 & 67.8 \\
                              & 0.90                  & 63.0 & 63.2 & 62.6 & 58.6 & 52.0 & 66.1 & 66.6 & 67.2 & 67.6 & 67.1 \\ \midrule
\multirow{5}{*}{OxfordPets}   & 0.50                  & 93.1 & 92.9 & 92.7 & 89.7 & 70.0 & 95.6 & 95.6 & 95.7 & 95.5 & 95.5 \\
                              & 0.75                  & 92.6 & 92.5 & 92.4 & 90.2 & 77.0 & 95.4 & 95.5 & 95.7 & 95.4 & 95.4 \\
                              & 0.80                  & 92.9 & 91.1 & 92.4 & 88.4 & 53.3 & 95.1 & 95.3 & 95.4 & 95.4 & 95.1 \\
                              & 0.88                  & 92.9 & 91.8 & 92.3 & 87.7 & 68.1 & 94.9 & 95.1 & 95.4 & 95.3 & 95.2 \\
                              & 0.90                  & 91.1 & 92.2 & 89.1 & 89.4 & 72.8 & 94.7 & 94.8 & 95.1 & 94.8 & 94.8 \\ \midrule
\multirow{5}{*}{DTD}          & 0.50                  & 82.9 & 82.9 & 81.5 & 80.3 & 80.4 & 80.5 & 81.0 & 80.2 & 80.8 & 81.3 \\
                              & 0.75                  & 83.0 & 82.6 & 81.4 & 82.0 & 74.1 & 79.4 & 79.5 & 80.1 & 80.7 & 80.8 \\
                              & 0.80                  & 82.3 & 82.2 & 82.5 & 81.9 & 81.1 & 77.9 & 78.3 & 79.3 & 79.5 & 79.4 \\
                              & 0.88                  & 74.7 & 78.3 & 79.6 & 79.4 & 76.4 & 70.2 & 71.7 & 72.3 & 73.2 & 73.8 \\
                              & 0.90                  & 76.3 & 77.5 & 76.5 & 77.8 & 76.8 & 68.5 & 69.2 & 70.6 & 72.3 & 72.9 \\ \midrule
\multirow{5}{*}{FGVCAircraft} & 0.50                  & 70.9 & 70.9 & 71.4 & 67.6 & 63.0 & 72.4 & 73.3 & 73.3 & 73.4 & 73.2 \\
                              & 0.75                  & 70.6 & 71.9 & 71.2 & 66.2 & 63.7 & 72.8 & 72.7 & 72.3 & 73.3 & 72.8 \\
                              & 0.80                  & 71.0 & 70.4 & 69.1 & 67.2 & 62.1 & 71.8 & 72.3 & 72.4 & 72.1 & 72.3 \\
                              & 0.88                  & 70.9 & 70.3 & 69.5 & 65.9 & 64.7 & 71.3 & 71.9 & 72.0 & 71.7 & 72.7 \\
                              & 0.90                  & 69.0 & 68.8 & 68.5 & 66.0 & 64.0 & 70.0 & 70.3 & 70.8 & 71.4 & 70.5 \\ \midrule
\end{tabular}}
\end{table}

\clearpage
\subsection{Comparison with Full-Data PLL Methods.} We conducted a systematic comparison with two recently proposed partial-label learning methods, Papi and CroSel. Unlike the 16-shot setting used in previous experiments, Papi and CroSel were trained on the entire datasets to fully exploit their optimal performance, while HopS was still trained under the 16-shot few-shot setting. The experiments were performed on two challenging datasets, Caltech and Oxford Flowers, both characterized by a large number of categories and high intra-class similarity. The results are shown in Table \ref{tab:threebackbones}.


\begin{table}[h]
  \centering
  \caption{Testing accuracy (\%) of HopS under \textit{uni}-prompt across three backbones.}
  \label{tab:threebackbones}
  \begin{tabular}{c||c||c|c|c||c|c|c||c|c|c}
    \toprule
    \multirow{2}{*}{Dataset} & \multirow{2}{*}{Method} & \multicolumn{3}{c||}{CLIP}  & \multicolumn{3}{c||}{R18} & \multicolumn{3}{c}{R50} \\ \cmidrule(lr){3-5} \cmidrule(lr){6-8} \cmidrule(lr){9-11}
                             &                         & 0.67 & 0.75 & 0.80 & 0.67 & 0.75 & 0.80 & 0.67 & 0.75 & 0.80 
                             \\ \midrule
    \multirow{3}{*}{Caltech} & HopS                    & 57.6 & \textbf{49.0} & \textbf{45.0} & \textbf{64.1} & \textbf{58.2} & \textbf{49.5} & \textbf{50.3} & \textbf{45.3} & \textbf{38.9} \\ 
                             & CroSel                  & \textbf{60.4} & 43.7 & 32.2 & 42.8 & 33.3 & 18.8 & 39.0 & 25.2 & 11.8 \\
                             & Papi                    & 59.7 & 49.2 & 40.5 & 44.3 & 35.2 & 15.8 & 42.0 & 23.5 & 13.4 \\ \midrule
    \multirow{3}{*}{Flowers} & HopS                    & \textbf{73.9} & \textbf{64.8} & \textbf{59.2} & \textbf{82.5} & \textbf{75.1} & \textbf{69.9} & \textbf{82.0} & \textbf{74.0} & \textbf{65.0} \\
                             & CroSel                  & 40.4 & 27.4 & 11.5 & 75.0 & 56.4 & 45.1 & 78.0 & 62.0 & 50.2 \\ 
                             & Papi                    & 35.7 & 13.7 & 5.8 & 67.9 & 56.1 & 47.0 & 67.0 & 58.7 & 49.8 \\ \bottomrule
  \end{tabular}
\end{table}

\subsection{Settings with Missing Ground-Truth.} We conduct 16-shot experiments on two types of label confusion, \textit{rand} and \textit{insd}, as previously described, with each candidate set \(S\) containing three labels. The missing rates of ground-truth labels are 12.5\% (2 out of 16 shots) for \textit{rand} and 25\% (4 out of 16 shots) for \textit{insd}. As shown in Table \ref{tab:4noise}, under the \textit{insd} setting, HopS consistently uncovers meaningful correlations within the candidate label sets, leading to significantly better performance than other methods. This suggests that HopS is particularly well-suited for real-world scenarios where label noise in SS is instance-dependent. In contrast, under the \textit{rand} setting where such dependencies are absent, HopS performs slightly worse than the SOTA robust loss. Nevertheless, its performance can be improved by integrating robust loss functions (e.g., the SCE loss). As illustrated in Table \ref{tab:2noise}, HopS combined with the SCE loss achieves the best performance under the \textit{rand} setting, demonstrating its flexibility and compatibility with other robust learning techniques.

\begin{table}[H]
\centering
\caption{Testing accuracy under the \textit{uni}-prompt with 25.0\% missing rate (\textit{insd}).}
\label{tab:4noise}
\begin{tabular}{@{\hskip 5pt}c||c|c|c|c|c|c|c|c|c|c@{\hskip 5pt}}
\toprule
Dataset & RC   & CC   & EXP  & LWC  & MAE  & MSE  & GCE  & SCE  & HopS          & HopS+SCE \\ \midrule
Caltech & 0.37 & 0.35 & 0.16 & 0.27 & 0.12 & 0.27 & 0.06 & 0.33 & \textbf{0.37} & 0.31     \\ \midrule
DTD     & 0.37 & 0.35 & 0.19 & 0.40 & 0.22 & 0.31 & 0.13 & 0.39 & \textbf{0.41} & 0.32     \\ \midrule
Flowers & 0.29 & 0.45 & 0.17 & 0.35 & 0.13 & 0.35 & 0.14 & 0.51 & \textbf{0.57} & 0.34     \\ \midrule
UCF     & 0.31 & 0.41 & 0.11 & 0.31 & 0.11 & 0.28 & 0.14 & 0.45 & \textbf{0.46} & 0.30     \\ \bottomrule
\end{tabular}
\end{table}

\begin{table}[H]
\centering
\caption{Testing accuracy under the \textit{uni}-prompt with 12.5\% missing rate (\textit{rand}).}
\label{tab:2noise}
\begin{tabular}{@{\hskip 5pt}c||c|c|c|c|c|c|c|c|c|c@{\hskip 5pt}}
\toprule
Dataset & RC   & CC   & EXP  & LWC  & MAE  & MSE  & GCE  & SCE  & HopS & HopS+SCE      \\ \midrule
Caltech & 0.76 & 0.84 & 0.34 & 0.84 & 0.38 & 0.20 & 0.26 & 0.86 & 0.84 & \textbf{0.87} \\ \midrule
DTD     & 0.52 & 0.51 & 0.19 & 0.53 & 0.23 & 0.19 & 0.15 & 0.54 & 0.53 & \textbf{0.55} \\ \midrule
Flowers & 0.53 & 0.87 & 0.18 & 0.73 & 0.18 & 0.13 & 0.15 & 0.87 & 0.88 & \textbf{0.89} \\ \midrule
UCF     & 0.50 & 0.63 & 0.18 & 0.57 & 0.14 & 0.08 & 0.10 & 0.64 & 0.64 & \textbf{0.64} \\ \bottomrule
\end{tabular}
\end{table}

\subsection{Running Time of Methods.} We measured the computational cost of each method during training, including the average epoch time (in seconds) and the total training time (in minutes), as shown in Table~\ref{tab:training_time}. Notably, although HopS introduces both a local module (LDF) and a global module (GOP) to enhance label identification, the additional computational overhead remains negligible due to the relatively small dataset size. As a result, HopS achieves superior performance without incurring a significant increase in training cost.
\begin{table}[ht]
\centering
\caption{Training efficiency comparison of different methods on Caltech.}
\label{tab:training_time}
\begin{tabular}{c||c|c|c|c|c|c|c|c|c}
\toprule
Metric & RC & CC & EXP & GCE & LWC & MAE & MSE & SCE & HopS \\
\midrule
Average Epoch Time (s) & 0.11 & 0.09 & 0.12 & 0.11 & 0.11 & 0.12 & 0.11 & 0.11 & 0.12 \\ \midrule
Total Time (min)       & 19.01 & 14.19 & 20.75 & 18.79 & 18.45 & 20.09 & 18.67 & 18.65 & 19.85 \\
\bottomrule
\end{tabular}
\end{table}

\subsection{Memory usage of large datasets.} As shown in Table \ref{tab:memory}, we compare the memory usage (in MiB) of HopS and Coop at different batch sizes on ImageNet, clearly showing that the memory cost of HopS is only slightly higher than that of Coop. HopS does not require an \(N^2\) dense affinity matrix. Instead, the kNN graph is computed sparsely for each batch, and the features come from a frozen encoder, allowing us to compute the kNN graph offline. Thus, HopS remains efficient and scalable even on non-few-shot PLL datasets, due to the sparse computation in its LDF module.

\begin{table}[h]
\centering
\caption{Memory usage comparison at different batch sizes on ImageNet}
\label{tab:memory}
\resizebox{0.8\linewidth}{!}{
\begin{tabular}{c|c|c|c|c|c|c|c|c}
\hline
Method & 16 & 32 & 64 & 128 & 256 & 512 & 1024 & 2048 \\ \hline
Coop & 18130 & 18102 & 18170 & 18312 & 19998 & 22178 & 19098 & 19564 \\ \hline
HopS & 18582 & 18582 & 18622 & 18764 & 20456 & 23312 & 19576 & 20430 \\ \hline
\(\triangle\) & + 452 & + 480 & + 452 & + 452 & + 458 & + 1134 & + 478 & + 866 \\ \hline
\end{tabular}}
\end{table}

\subsection{Sensitivity to \(k\) and \(\tau\).} We evaluate the sensitivity of LDF by sweeping $k \in \{10, 20, 40\}$ (with $\tau = 0.4$) and $\tau \in \{0.4, 0.5, 0.6\}$ (with \(k = 20\)) on coarse/fine-grained datasets (Caltech/Flowers), under \textit{rand} and \textit{insd} confusion types, using both \textit{uni}-prompt (left) and \textit{cls}-prompts (right). As shown in Table \ref{tab:sens}, performance varies smoothly with \(k\) and \(\tau\), with no significant re-tuning needed from Caltech to Flowers.

\begin{table}[h]
\centering
\caption{Sensitivity of LDF to \(k\) and \(\tau\).}
\label{tab:sens}
\resizebox{1.\linewidth}{!}{
\begin{tabular}{c||c||c|c|c|c|c||c|c|c|c|c}
\hline
{Dataset} & {Type} & {k=10} & {k=20} & {k=40} & {\(\tau\)=0.5} & {\(\tau\)=0.6} & {k=10} & {k=20} & {k=40} & {\(\tau\)=0.5} & {\(\tau\)=0.6} \\ \hline
\multirow{4}{*}{Caltech} & CLIP & 48.9 & 49.0 & 45.6 & 56.3 & 53.1 & 52.4 & 60.0 & 48.8 & 56.8 & 54.0 \\
 & R18  & 45.3 & 58.2 & 47.9 & 51.4 & 49.9 & 54.5 & 59.4 & 53.1 & 65.0 & 62.4 \\
 & R50  & 41.7 & 45.3 & 40.0 & 47.3 & 48.8 & 59.9 & 53.3 & 49.3 & 50.1 & 54.1 \\
 & Rand  & 89.4 & 89.0 & 88.4 & 89.4 & 89.0 & 90.0 & 90.4 & 90.1 & 90.1 & 90.0 \\
\hline
\multirow{4}{*}{Flowers} & CLIP & 63.1 & 64.8 & 60.6 & 65.8 & 66.0 & 72.4 & 68.8 & 61.8 & 69.6 & 68.8 \\
 & R18  & 73.4 & 75.1 & 74.0 & 78.1 & 76.2 & 83.8 & 80.4 & 74.6 & 80.0 & 79.9 \\
 & R50  & 76.2 & 74.0 & 80.1 & 73.5 & 74.4 & 80.5 & 78.2 & 81.0 & 78.1 & 77.9 \\
 & Rand  & 91.8 & 92.9 & 92.5 & 91.9 & 92.0 & 95.5 & 95.7 & 95.5 & 95.6 & 95.6 \\
\hline
\end{tabular}}
\end{table}

\subsection{Sensitivity to \(\varepsilon\).} We evaluate the sensitivity of GOP by sweeping OT entropy coefficient \(\varepsilon\) on coarse- and fine-grained datasets, i.e., Caltech and Flowers, under both RAND and INSD confusions, using both \textit{uni}-prompt and \textit{cls}-prompts. As shown in Table~\ref{tab:epsc} and ~\ref{tab:epsf}, our method remains effective under different \(\varepsilon\), and \(\varepsilon=0.05\) lies in a strong and stable range.
\begin{table}[ht]
\centering
\caption{Sensitivity analysis of \(\varepsilon\) on Caltech.}
\label{tab:epsc}
\begin{tabular}{c||c|c|c|c|c|c|c|c|c}
\toprule
Type & 0.02 & 0.05 & 0.08 & 0.10 & 0.50 & 1 & 2 & 5 & 10 \\
\midrule
\textit{uni}-RAND  & 88.8 & 89.0 & 88.7 & \textbf{89.5} & 88.6 & 88.0 & 89.3 & 88.6 & 88.2 \\
\textit{cls}-RAND & \textbf{90.4} & \textbf{90.4} & 87.7 & 87.6 & 87.4 & 87.5 & 87.4 & 87.4 & 87.8 \\
\midrule 
\textit{uni}-INSD  & 45.2 & \textbf{49.0} & 43.9 & 43.7 & 39.0 & 43.6 & 41.1 & 42.3 & 41.3 \\
\textit{cls}-INSD & 53.0 & \textbf{60.0} & 42.5 & 47.5 & 40.7 & 41.4 & 39.8 & 41.2 & 45.1 \\
\bottomrule
\end{tabular}

\end{table}
\begin{table}[ht]
\centering
\caption{Sensitivity analysis of \(\varepsilon\) on Flowers.}
\label{tab:epsf}
\begin{tabular}{c||c|c|c|c|c|c|c|c|c}
\toprule
Type & 0.02 & 0.05 & 0.08 & 0.10 & 0.50 & 1 & 2 & 5 & 10 \\
\midrule
\textit{uni}-RAND  & 91.8 & \textbf{92.9} & 91.5 & 91.2 & 91.2 & 91.6 & 91.1 & 90.9 & 90.2 \\
\textit{cls}-RAND  & 95.5 & \textbf{95.7} & 94.1 & 93.6 & 93.6 & 93.4 & 93.7 & 93.7 & 93.5 \\
\midrule 
\textit{uni}-INSD  & 62.4 & \textbf{64.8} & 64.6 & 59.4 & 62.3 & 61.6 & 62.6 & 62.0 & 62.5 \\
\textit{cls}-INSD  & \textbf{69.5} & 68.8 & 64.6 & 63.0 & 64.8 & 64.3 & 63.3 & 63.8 & 65.3 \\
\bottomrule
\end{tabular}
\end{table}

\subsection{Settings of 4/8-Shot.} 
\begin{figure}[h]
    \centering
    \begin{minipage}{\linewidth}
        \centering
        \includegraphics[width=0.95\linewidth]{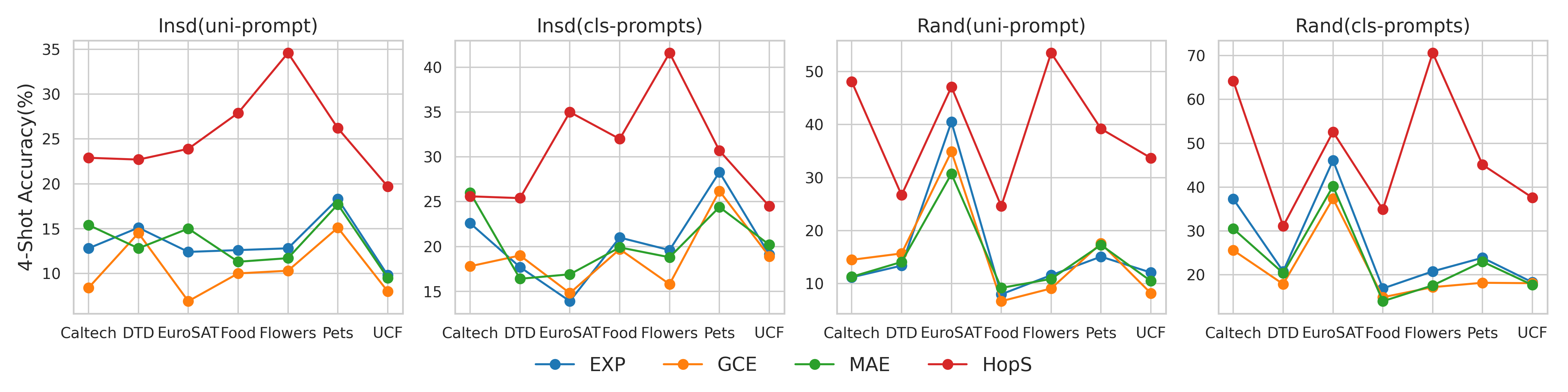}
    \end{minipage}
    \begin{minipage}{\linewidth}
        \centering
        \includegraphics[width=0.95\linewidth]{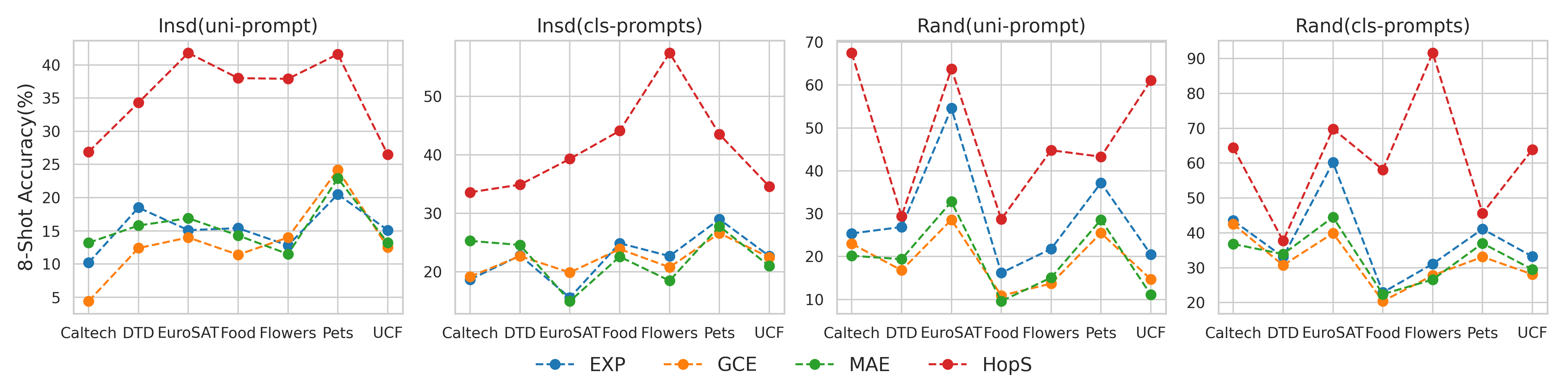}
    \end{minipage}
    \caption{Comparison results under 4/8-shot settings across eight datasets}
    \label{fig:shot}
\end{figure}

As clarified in Figure~\ref{fig:shot}, we additionally conducted 4-shot and 8-shot experiments to evaluate HopS under more limited supervision. The results show that HopS consistently outperforms the baselines across all eight datasets, further demonstrating its effectiveness in low-shot partial-label learning scenarios.

\end{document}